\documentclass{article}

\PassOptionsToPackage{table}{xcolor}
\usepackage{xcolor}
\usepackage{iclr2027_conference,times}
\usepackage{latexsym}
\usepackage[T1]{fontenc}
\usepackage[utf8]{inputenc}
\usepackage{microtype}
\usepackage{inconsolata}
\usepackage{hyperref}
\usepackage{url}
\usepackage{tikz}
\usepackage{pgfplots}
\usepgfplotslibrary{groupplots}
\usetikzlibrary{shapes.geometric}
\pgfplotsset{compat=1.18}
\usepackage{booktabs}
\usepackage{amsmath,amssymb}
\usepackage{graphicx}
\usepackage{array}
\usepackage{tcolorbox}
\usepackage{adjustbox}
\usepackage{wrapfig}
\tcbuselibrary{breakable}

\newtcolorbox{promptbox}[1]{
    breakable,
    colback=gray!4,
    colframe=gray!45,
    colbacktitle=gray!12,
    coltitle=black,
    boxrule=0.5pt,
    arc=1mm,
    title={#1},
    fonttitle=\bfseries\small,
    left=2mm,
    right=2mm,
    top=1.5mm,
    bottom=1.5mm,
    before upper={\small\ttfamily}
}
\usepackage{tabularx}
\usepackage{ragged2e}
\usepackage{multirow}
\usepackage{makecell}
\usepackage{enumitem}
\pgfplotsset{compat=1.18}
\usepackage{float}
\usepackage{capt-of}
\newcolumntype{L}{>{\RaggedRight\arraybackslash}X}
\newcommand{\Buzzy}{\textsc{Buzzy}}
\newcommand{\method}{\textsc{Buzzy}}
\newcommand{\mm}{\mathrm{mm}}
\newcommand{\txt}{\mathrm{txt}}

\newcommand{\argmax}{\mathop{\mathrm{argmax}}}

\title{BUZZY: Contrastive Scoring to Mitigate \\ Text-Induced Bias in Multimodal Multiple-Choice QA}

\author{
\parbox{\textwidth}{
\raggedright
\textbf{Taeyun Roh}$^{1}$, \textbf{Suhyeong Park}$^{1,4}$, \textbf{Dongyoung Lee}$^{1}$, \textbf{Wonjune Jang}$^{3,4}$, \\ 
\textbf{Junha Jung}$^{1,4}$, \textbf{EunYeong Jo}$^{2}$, \textbf{Jaewoo Kang}$^{1,4}$ \\[6pt]
{\normalfont \raggedright % 정렬이 풀리지 않도록 한 번 더 명시
$^{1}$Korea University \quad
$^{2}$KAIST \quad
$^{3}$Myongji University \quad
$^{4}$AIGEN Sciences \\
\texttt{\{nrbsld,kangj\}@korea.ac.kr}
}
}
}
\definecolor{linkpink}{RGB}{219, 70, 128}
\definecolor{citeblue}{RGB}{45, 90, 180}

\hypersetup{
  pdftitle={BUZZY: Contrastive Scoring to Mitigate Text-Induced Bias in Multimodal Multiple-Choice QA},
  pdfauthor={Taeyun Roh, Suhyeong Park, Dongyoung Lee, WonJune Jang, Junha Jung, EunYeong Jo, Jaewoo Kang},
  colorlinks=true,
  linkcolor=black,      % Figure/Table/Section references
  citecolor=citeblue,   % citations
  urlcolor=linkpink     % external URLs / project page
}

\iclrfinalcopy
\begin{document}

\maketitle
% arXiv/preprint copy: suppress ICLR review/publication header.

\begin{abstract}
Multimodal multiple-choice question answering (MCQA) provides a standardized and objectively measurable setting for evaluating vision-language models (VLMs). However, because the MCQA format incorporates the candidate choices into the input context, it introduces several unintended biases. Previous work has primarily focused on structural biases, such as preferences for certain position choices. Instead, we argue that the choices act as textual priors, causing models to favor linguistically plausible options regardless of the visual content. We hypothesize and empirically verify that a model genuinely relies on visual evidence only when its multimodal distribution significantly diverges from its text-only distribution. Based on this observation we propose \method{}, a training-free decoding method that corrects multimodal predictions by subtracting the text-only distribution. Experiments with five VLMs on five multimodal MCQA benchmarks demonstrate that \method{} achieves the highest average accuracy among state-of-the-art methods while reducing inference latency by over 28\% compared to prior contrastive scoring approaches. Overall, these results suggest that amplifying the visual signal by penalizing text-only preferences is key to efficient and robust multimodal MCQA reasoning. Code and additional resources are provided
\href{https://txxnrd.github.io/buzzy/}{here}.
% Project-page link removed for anonymous ICLR 2027 review submission.
\end{abstract}

\section{Introduction}
Vision-language models (VLMs) have demonstrated strong capabilities across a wide range of tasks~\citep{li2023blip, liu2024improved,jung2026breaking}. Multimodal multiple-choice question answering (MCQA) has consequently become a standard setting for evaluating whether these models can reason over visual evidence~\citep{liu2024mmbench, yue2024mmmu}. Compared with open-ended generation, MCQA enables objective and reproducible evaluation by avoiding ambiguities in answer matching~\citep{pmlr-v250-royer24a,zhang2025automated}. Its structured answer space, however, also changes how model predictions should be interpreted~\citep{balepur2024artifactsabductionllmsanswer, atabuzzaman2025benchmarkingmitigatingmcqaselection}.

% However, because the MCQA format inherently incorporates candidate choices directly into the input context, it introduces critical unintended biases. Previous research has primarily addressed structural biases, such as positional preferences for certain option letters, developing calibration techniques to re-balance prediction probabilities. Despite these efforts, we highlight a more fundamental flaw: the candidate choices themselves act as strong textual priors. This leads VLMs to favor linguistically plausible options based solely on language association, bypassing visual evidence entirely.

In multimodal MCQA, candidate choices are not merely possible outputs. They are presented to the model as part of the textual input before prediction. 
Therefore, choices expose the model to textual priors that favor linguistically plausible answers regardless of the image, and structural biases that create preferences for particular choice positions~\citep{balepur2024artifactsabductionllmsanswer, luo2025probing, pezeshkpour2024large, yang2025option}. High multimodal confidence therefore does not necessarily indicate strong visual grounding because it reflects both visual evidence and preferences induced by the question and choices~\citep{atabuzzaman2025benchmarkingmitigatingmcqaselection,loginova-etal-2025-addressing}. To empirically test this, we conduct a pilot study comparing each model's multimodal and text-only answer distributions. We find that correct predictions deviate substantially from the text-only distribution, whereas incorrect predictions remain close. Thus, authentic visual reasoning is best measured not by absolute confidence, but by the extent of deviation from textual baselines.

%In multimodal MCQA, candidate choices are not merely possible outputs. They are presented to the model as part of the textual input before prediction. 
%Therefore, choices expose the model to both textual priors that favor linguistically plausible answers regardless of the image, and structural biases that create preferences for particular choice positions ~\citep{balepur2024artifactsabductionllmsanswer, luo2025probing}. High multimodal confidence therefore does not necessarily indicate strong visual grounding because it reflects both visual evidence and preferences induced by the question and choices. To empirically test this, we conduct a pilot study comparing each model's multimodal and text-only answer distributions. We find that correct predictions deviate substantially from the text-only distribution, whereas incorrect predictions remain close. Thus, authentic visual reasoning is best measured not by absolute confidence, but by the extent of deviation from textual baselines.

Contrastive decoding is a widely adopted framework for mitigating these biased distributions, effectively separating useful visual evidence from undesirable model preferences~\citep{obrien2023contrastive}. Methods such as Visual Contrastive Decoding (VCD)~\citep{leng2024vcd} and Instruction Contrastive Decoding (ICD)~\citep{wang2024icd} contrast standard predictions with those obtained from distorted images or disturbance instructions. 
% in open-ended generation. 
However, these methods have two main limitations. First, their distorted contrastive inputs do not directly capture the textual prior introduced by the choices. Second, they require two multimodal forward passes, repeatedly processing the visual tokens and introducing inference overhead.

We introduce \textbf{\method{}}, a training-free decoding method built on a simple principle: a candidate should be rewarded for the support it gains from the image, not for the plausibility it already possesses from text alone. The name \method{} reflects the idea of suppressing the ``buzz'' of text-induced priors, so that visually grounded evidence becomes clearer. Rather than constructing a distorted negative input, \method{} uses the text-only context as a natural control condition. It obtains candidate logits from multimodal and text-only inputs and subtracts the text-only logits from their multimodal counterparts. The resulting score discounts preferences explainable by language alone and emphasizes the features grounded in the visual context.

Across five VLM backbones and five multimodal MCQA benchmarks, \method{} consistently outperforms greedy, prompt-based, and contrastive decoding baselines. Averaged across all backbones and benchmarks, \method{} achieves 57.84 accuracy and 53.56 Macro-F1, exceeding the strongest overall baseline by 1.99 and 1.85 points, respectively. It also reduces inference latency by over 28\% compared with prior multimodal contrastive decoding methods. 

% These results demonstrate that measuring candidate support relative to text-only preference provides an effective and efficient approach to visually grounded MCQA.

Our contributions are as follows:
\begin{itemize}
    \item We characterize visual grounding through the discrepancy between multimodal and text-only preferences. We show that incorrect predictions exhibit greater similarity to text-only preferences than correct predictions, revealing the bias induced by choices.
    \item We propose \method{}, a training-free contrastive scoring method that uses the text-only context as a natural control and prioritizes candidate support that cannot be explained by textual preference alone.
    \item We show that \method{} achieves the highest average performance across five benchmarks and five VLM backbones while reducing inference latency by over 28\% compared with prior multimodal contrastive decoding methods.
\end{itemize}

\section{Related work}

\subsection{Choice Bias in MCQA}
Prior work has identified several forms of bias in MCQA, including positional preferences, option-order sensitivity, and answer-label bias~\citep{pezeshkpour2024large,yang2025option}. Models can also exploit semantic or stylistic artifacts within the candidate set and sometimes answer above chance using the choices alone~\citep{wang-etal-2023-dataset,balepur2024artifactsabductionllmsanswer,wang2024llmsperformmcqaselecting,zheng2024largelanguagemodelsrobust,atabuzzaman2025benchmarkingmitigatingmcqaselection}. Calibration, answer permutation, and related inference-time strategies have been proposed to reduce sensitivity to these artifacts~\citep{zheng2024largelanguagemodelsrobust,atabuzzaman2025benchmarkingmitigatingmcqaselection,loginova-etal-2025-addressing}. As a representative example, Adaptive Logit Correction (ALC) reduces option-label and positional biases by calibrating and correcting multimodal logits~\citep{atabuzzaman2025benchmarkingmitigatingmcqaselection}. However, prior work has largely concentrated on structural biases, treating textual answer-space bias and visual grounding as entirely separate problems. We connect the two by demonstrating that candidate-induced textual preferences persist in multimodal predictions. 
% To address this, we use text-only scores to explicitly discount these priors during answer selection.
 
\subsection{Contrastive Decoding for Multimodal Grounding}

Contrastive decoding steers model predictions by comparing a target distribution against a less desirable reference distribution~\citep{obrien2023contrastive}. In multimodal settings, this reference is commonly constructed by perturbing the visual input or instruction~\citep{lee2024delve,park2025convis}. Representative methods include VCD, which contrasts predictions from original and distorted images~\citep{leng2024vcd}, and ICD, which uses a disturbance instruction for its contrastive branch~\citep{wang2024icd}. Subsequent methods dynamically select or weight contrastive signals to address hallucination and cross-modal knowledge conflicts~\citep{zhu2024unraveling,suo2025octopus}.

\section{Preliminaries}

To distinguish genuine visual grounding from textual bias, we formalize the MCQA setting and compare answer distributions with and without images.

\subsection{Problem Setting}
\label{subsec:problem_setting}

Each multimodal MCQA example consists of an image $x$, a question $q$, and a set of candidate answers $\mathcal{C}=\{c_1,\dots,c_K\}$. The goal is to predict the correct answer $y^*\in\mathcal{C}$. Given a VLM, we define two distributions over the same candidate set. The multimodal distribution $p_{\mathrm{mm}}(c\mid x,q,\mathcal{C})$ is obtained from the complete input, whereas the text-only distribution $p_{\mathrm{txt}}(c\mid q,\mathcal{C})$ is obtained by removing only the image while preserving the question and choices. The former reflects the model's preference when visual and textual information are available, while the latter estimates the preference induced by text alone. To make the distributions directly comparable, we require the model to output a single option-label token without generating intermediate reasoning.

% \subsection{Pilot Study}
\subsection{Motivating Analysis}
\label{subsec:pilot_study}
\begin{figure}[t]
    \centering
    \includegraphics[width=0.92\linewidth]{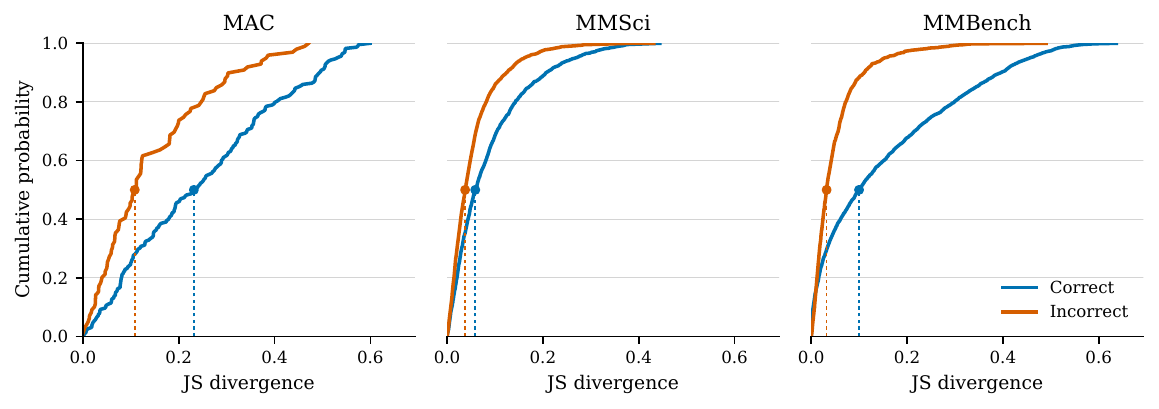}
    \caption{
    Empirical cumulative distribution functions (ECDFs) of the
    JS divergence between multimodal and text-only answer distributions
    for correct and incorrect predictions across three benchmarks.
    Correct predictions generally exhibit larger JS divergence,
    indicating greater deviation from text-only preferences.
    }
    \label{fig:motivation_js_ecdf}
\end{figure}

 We compute $p_{\mathrm{mm}}(c\mid x,q,\mathcal{C})$ and $p_{\mathrm{txt}}(c\mid q,\mathcal{C})$ under greedy decoding, take the multimodal prediction $\hat{y}=\argmax_{c\in\mathcal{C}}p_{\mathrm{mm}}(c\mid x,q,\mathcal{C})$, and partition the examples according to whether $\hat{y}=y^*$. We then compare the two distributions using Jensen--Shannon (JS) divergence~\citep{lin1991divergence} which quantifies the statistical distance between the distributions. As shown in Figure~\ref{fig:motivation_js_ecdf}, correct predictions
are consistently shifted toward higher JS divergence than incorrect
predictions. The median divergence at a cumulative probability of 0.5
is also higher for correct predictions across all three benchmarks.
This indicates that correct predictions tend to deviate more strongly
from the text-only distribution. The remaining two benchmarks show the
same overall trend, as reported in Appendix~\ref{sec:additional_js}.

\begin{figure*}[t] % [t]는 페이지 상단(top) 배치를 의미합니다.
    \centering
    % \textwidth를 0.85\textwidth로 수정하여 크기를 줄였습니다. (비율은 조절 가능)
    \includegraphics[width=0.90\textwidth, keepaspectratio]{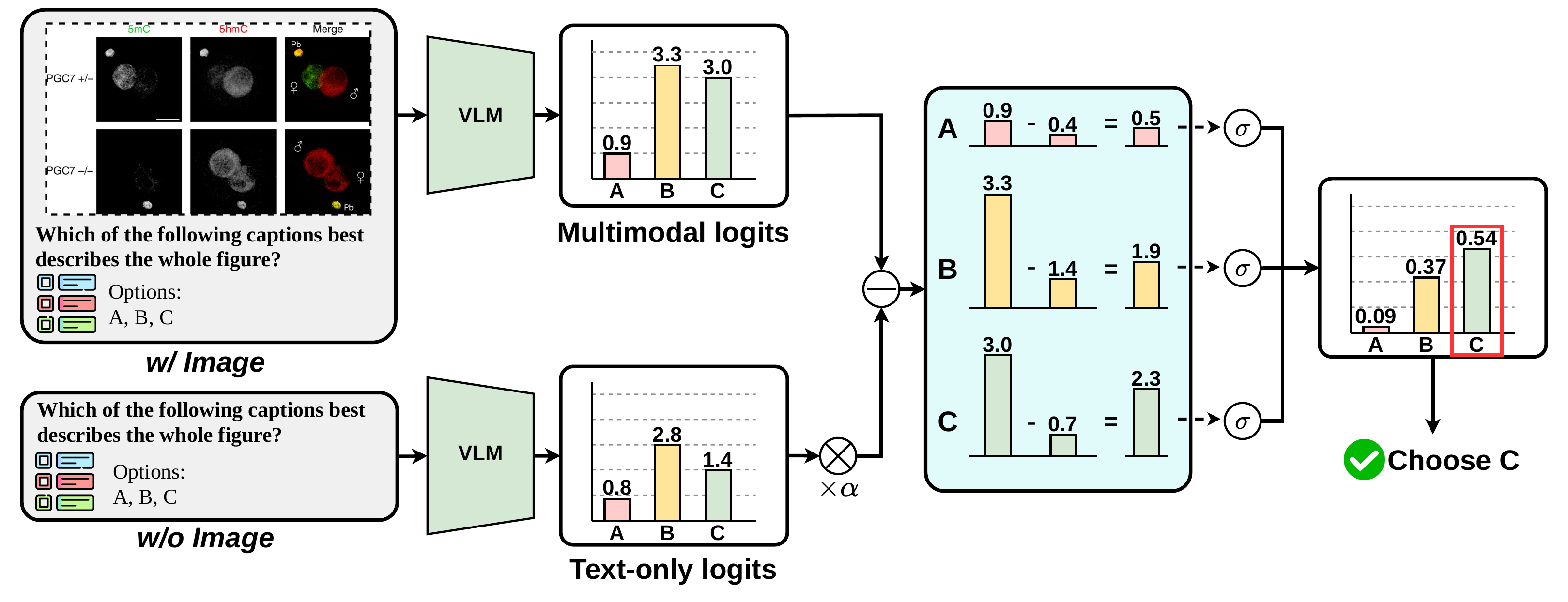}
    \caption{Illustration of \Buzzy. Given a question and candidate answers, the model produces candidate scores under both multimodal and text-only inputs. \Buzzy\ subtracts the text-only score, scaled by $\alpha$, from the multimodal score so that candidates favored mainly by textual prior are suppressed and visually grounded candidates are promoted.}
    \label{fig:main_method}
\end{figure*}
\section{Method}
\label{sec:method}

% Motivated by the pilot study, we propose \Buzzy, a training-free decoding method that suppresses text-induced preferences in multimodal MCQA. \Buzzy\ consists of three stages: it first obtains candidate logits from multimodal and text-only contexts, then removes the text-only contribution through contrastive correction, and finally selects the answer using the corrected scores. This procedure ensures that the selected answer is driven by genuine visual evidence. Figure~\ref{fig:main_method} illustrates the overall decoding procedure.

% Motivated by the pilot study, we propose \Buzzy, a training-free decoding method that suppresses text-induced preferences in multimodal MCQA. \Buzzy\ first obtains candidate logits from matched multimodal and text-only contexts. It then discounts the text-only preference through contrastive correction and selects the answer using the corrected scores. This procedure favors candidate support that cannot be explained by textual information alone. Figure~\ref{fig:main_method} illustrates the overall decoding procedure.

Motivated by the pilot study, we propose \Buzzy, a training-free decoding method that suppresses text-induced preferences in multimodal MCQA. \Buzzy\ consists of three stages. First, it obtains candidate logits from matched multimodal and text-only contexts, using the latter as a proxy for preferences induced by the question and choices alone. Second, it subtracts the weighted text-only logits from their multimodal counterparts to discount candidate support explainable without the image. Finally, it normalizes the corrected logits and selects the highest-scoring candidate as the final answer. Figure~\ref{fig:main_method} illustrates the overall decoding procedure.

\subsection{Dual-Context Candidate Scoring}
\label{subsec:dual_context}

For each question, we evaluate the model under the two contexts defined in Section~\ref{subsec:problem_setting}. The multimodal context contains the image $x$, question $q$, and candidate set $\mathcal{C}$, whereas the text-only context removes $x$ while preserving the question and candidate set.  
% Controlling the textual components ensures that the difference between the two branches primarily reflects the contribution of visual information. 
Fixing the textual components ensures that visual signals remain the only distinguishing factor between the input contexts. The two contexts produce candidate-wise logits over the same answer set:
\begin{align}
l_{\mm}(c)
    &= \mathrm{logit}_{\theta}(c\mid x,q,\mathcal{C}), \\[4pt]
l_{\txt}(c)
    &= \mathrm{logit}_{\theta}(c\mid q,\mathcal{C}).
\end{align}

Here, $\mathrm{logit}_{\theta}$ denotes the pre-softmax score assigned by the language modeling head to the token corresponding to candidate $c$, such as the option label A or B. The multimodal logit $l_{\mm}(c)$ reflects the model's preference when visual and textual evidence are available, whereas $l_{\txt}(c)$ captures the preference induced without access to the image. 
We use the text-only branch not as an independent predictor, but as a proxy for the textual preference that may influence the multimodal prediction.

\subsection{Text-Prior Contrastive Correction}
\label{subsec:contrastive_correction}

Although the multimodal branch receives the image, its prediction may still be dominated by a candidate that appears plausible from the textual context alone. To reduce this effect, \Buzzy\ explicitly discounts the text-only logit from the corresponding multimodal logit:
\begin{equation}
l_{\mathrm{buzzy}}(c)
=
\underbrace{l_{\mm}(c)}_{\text{multimodal score}}
-
\underbrace{\alpha l_{\txt}(c)}_{\text{text-prior penalty}}
\label{eq:Buzzy}
\end{equation}
where $\alpha\geq0$ controls the strength of the penalty. Setting $\alpha=0$ recovers standard greedy decoding, while increasing $\alpha$ applies a stronger penalty to candidates favored by the text-only branch.

This correction suppresses candidates whose multimodal scores can largely be explained by text-only preferences. A linguistically plausible distractor favored by both branches receives a larger penalty, whereas a candidate supported primarily by the image retains a relatively high corrected score. 
% \Buzzy\ therefore promotes visually grounded evidence while directly addressing the strong alignment between incorrect multimodal predictions and text-only preferences observed in our pilot study.
\Buzzy\ therefore favors candidate support that cannot be explained by text alone, directly addressing the alignment between incorrect multimodal predictions and text-only preferences observed in our pilot study.

\subsection{Final Prediction}
\label{subsec:final_prediction}

After applying the contrastive correction, we normalize the corrected logits and select the most probable candidate:
\begin{equation}
\begin{aligned}
p_{\mathrm{buzzy}}(c)
&=
\frac{\exp(l_{\mathrm{buzzy}}(c))}
{\sum_{c'\in\mathcal{C}}\exp(l_{\mathrm{buzzy}}(c'))},
\\[4pt]
\hat{y}
&=
\argmax_{c\in\mathcal{C}}p_{\mathrm{buzzy}}(c).
\end{aligned}
\label{eq:final_prediction}
\end{equation}
Because softmax preserves the ranking of the corrected logits, this prediction is equivalent to directly maximizing $l_{\mathrm{buzzy}}(c)$, allowing normalization to be omitted when only the final answer is required.

\Buzzy\ requires one standard multimodal forward pass and one additional text-only forward pass. It introduces no parameter updates, auxiliary models, or changes to the model architecture, making it directly applicable to VLMs that expose candidate-level logits. Moreover, because the second pass does not process visual tokens, it incurs substantially lower computational overhead than contrastive decoding methods that require two complete multimodal passes.
\begin{table*}[t]
\centering
\footnotesize
\setlength{\tabcolsep}{2.4pt}
\renewcommand{\arraystretch}{0.94}

\resizebox{0.96\textwidth}{!}{%
\begin{tabular}{ll cc cc cc cc cc cc}
\toprule
\multirow{2}{*}{\textbf{Backbone}}
& \multirow{2}{*}{\textbf{Method}}
& \multicolumn{2}{c}{\textbf{MAC}}
& \multicolumn{2}{c}{\textbf{SciFIBench}}
& \multicolumn{2}{c}{\textbf{MMSci}}
& \multicolumn{2}{c}{\textbf{MMBench}}
& \multicolumn{2}{c}{\textbf{MMStar}}
& \multicolumn{2}{c}{\textbf{Avg.}} \\

\cmidrule(lr){3-4}
\cmidrule(lr){5-6}
\cmidrule(lr){7-8}
\cmidrule(lr){9-10}
\cmidrule(lr){11-12}
\cmidrule(lr){13-14}

& & ACC & F1
& ACC & F1
& ACC & F1
& ACC & F1
& ACC & F1
& \textbf{ACC} & \textbf{F1} \\
\midrule

% ---------------------------------------------------------
% Qwen 3.5 4B
% ---------------------------------------------------------
\multirow{6}{*}{Qwen 3.5 4B}
& Greedy
& 69.72 & 70.75
& 46.20 & 44.93
& 38.83 & 20.49
& 74.38 & 74.43
& 38.13 & 36.62
& 53.45 & 49.44 \\

& Prompt-enhanced
& 63.00 & 62.43
& 42.00 & 39.96
& 35.84 & 20.30
& 69.81 & 69.73
& 36.47 & 34.51
& 49.42 & 45.39 \\

& VCD
& 68.50 & 68.77
& 45.40 & 44.00
& 40.85 & \textbf{26.58}
& 75.35 & 75.40
& 38.93 & 37.29
& 53.81 & 50.41 \\

& ICD
& 58.41 & 58.57
& 40.50 & 38.23
& 33.79 & 14.83
& 70.48 & 70.45
& 35.27 & 33.60
& 47.69 & 43.14 \\

& ALC
& 68.81 & 68.93
& 47.50 & 46.36
& 39.69 & 19.05
& 74.61 & 74.64
& 38.67 & 37.51
& 53.86 & 49.30 \\

\rowcolor{blue!10}
\cellcolor{white}
& \textbf{\Buzzy\ (Ours)}
& \textbf{74.01} & \textbf{74.24}
& \textbf{48.70} & \textbf{46.99}
& \textbf{43.44} & 19.06
& \textbf{76.16} & \textbf{76.17}
& \textbf{40.27} & \textbf{38.02}
& \textbf{56.52} & \textbf{50.90} \\

\midrule

% ---------------------------------------------------------
% Qwen 3.5 9B
% ---------------------------------------------------------
\multirow{6}{*}{Qwen 3.5 9B}
& Greedy
& 81.35 & 81.40
& 55.10 & 54.67
& 46.54 & 29.05
& 76.48 & 76.43
& 41.00 & 40.59
& 60.09 & 56.43 \\

& Prompt-enhanced
& 81.65 & 81.69
& 54.10 & 53.65
& 45.22 & 23.53
& 77.64 & 77.63
& 40.47 & 40.48
& 59.82 & 55.40 \\

& VCD
& 81.96 & 81.92
& 55.90 & 55.55
& 49.29 & 32.35
& 78.10 & 78.09
& 40.87 & 40.81
& 61.22 & 57.74 \\

& ICD
& 81.04 & 81.01
& 53.00 & 52.13
& 46.65 & 24.14
& 72.88 & 72.84
& 39.87 & 39.63
& 58.69 & 53.95 \\

& ALC
& 78.59 & 78.49
& 53.10 & 52.87
& 46.78 & 27.07
& 76.48 & 76.41
& 40.47 & 39.96
& 59.08 & 54.96 \\

\rowcolor{blue!10}
\cellcolor{white}
& \textbf{\Buzzy\ (Ours)}
& \textbf{82.26} & \textbf{82.34}
& \textbf{58.00} & \textbf{57.55}
& \textbf{52.14} & \textbf{33.91}
& \textbf{79.14} & \textbf{79.08}
& \textbf{43.73} & \textbf{43.39}
& \textbf{63.05} & \textbf{59.25} \\

\midrule

% ---------------------------------------------------------
% Phi-3.5
% ---------------------------------------------------------
\multirow{6}{*}{
\begin{tabular}[c]{@{}l@{}}
Phi-3.5-\\
vision-instruct
\end{tabular}}
& Greedy
& 42.81 & 42.02
& 48.60 & 48.47
& 47.78 & 34.43
& 79.23 & 79.44
& 45.33 & 42.89
& 52.75 & 49.45 \\

& Prompt-enhanced
& 45.26 & 44.64
& 49.10 & 49.01
& 48.21 & 33.58
& 78.91 & 79.09
& 47.13 & 45.02
& 53.72 & 50.27 \\

& VCD
& 43.73 & 43.27
& 53.50 & 53.75
& 51.95 & \textbf{34.59}
& 79.76 & 79.87
& 47.87 & 46.47
& 55.36 & 51.59 \\

& ICD
& 42.81 & 41.60
& 47.10 & 46.86
& 46.32 & 31.38
& 77.89 & 78.16
& 45.20 & 42.80
& 51.86 & 48.16 \\

& ALC
& 44.65 & 44.32
& 47.60 & 47.28
& 47.29 & 30.84
& 78.93 & 79.04
& 45.73 & 43.93
& 52.84 & 49.08 \\

\rowcolor{blue!10}
\cellcolor{white}
& \textbf{\Buzzy\ (Ours)}
& \textbf{49.54} & \textbf{49.89}
& \textbf{54.90} & \textbf{55.02}
& \textbf{52.71} & 29.93
& \textbf{80.18} & \textbf{80.22}
& \textbf{48.13} & \textbf{47.10}
& \textbf{57.09} & \textbf{52.43} \\

\midrule

% ---------------------------------------------------------
% Molmo
% ---------------------------------------------------------
\multirow{6}{*}{Molmo-7B-D}
& Greedy
& 36.39 & 33.87
& 40.10 & 39.69
& 30.48 & 8.16
& 72.53 & 72.73
& 41.00 & 35.52
& 44.10 & 37.99 \\

& Prompt-enhanced
& 41.59 & 40.52
& 42.30 & 41.49
& 31.61 & 8.48
& 75.12 & 75.18
& \textbf{43.40} & 38.67
& 46.80 & 40.87 \\

& VCD
& 40.06 & 38.21
& \textbf{42.50} & \textbf{42.25}
& 31.96 & 14.02
& 74.87 & 74.93
& 42.87 & 38.97
& 46.45 & 41.68 \\

& ICD
& 35.78 & 32.72
& 36.80 & 35.41
& 27.92 & 5.39
& 68.31 & 68.30
& 39.47 & 32.22
& 41.66 & 34.81 \\

& ALC
& 36.09 & 33.90
& 41.30 & 40.92
& 30.32 & 8.34
& 73.46 & 73.59
& 41.53 & 36.18
& 44.54 & 38.59 \\

\rowcolor{blue!10}
\cellcolor{white}
& \textbf{\Buzzy\ (Ours)}
& \textbf{43.12} & \textbf{42.00}
& 40.90 & 40.57
& \textbf{32.26} & \textbf{18.73}
& \textbf{75.24} & \textbf{75.41}
& 43.13 & \textbf{39.33}
& \textbf{46.93} & \textbf{43.21} \\

\midrule

% ---------------------------------------------------------
% GLM
% ---------------------------------------------------------
\multirow{6}{*}{GLM-4.6V-Flash}
& Greedy
& 73.70 & 73.75
& 60.30 & 60.28
& 51.20 & 32.48
& 79.33 & 79.37
& 44.80 & 43.16
& 61.87 & 57.81 \\

& Prompt-enhanced
& 71.87 & 71.86
& 59.50 & 59.58
& 50.15 & 22.83
& 76.90 & 76.71
& 42.40 & 39.96
& 60.16 & 54.19 \\

& VCD
& 71.25 & 71.36
& \textbf{61.90} & \textbf{61.88}
& 52.92 & 27.87
& 80.62 & 80.71
& 45.33 & 43.70
& 62.40 & 57.10 \\

& ICD
& 67.89 & 67.85
& 53.80 & 53.42
& 45.97 & 24.57
& 78.03 & 78.15
& 43.93 & 42.92
& 57.92 & 53.38 \\

& ALC
& 70.03 & 70.02
& 60.20 & 60.17
& 49.39 & 26.68
& 79.42 & 79.49
& 44.67 & 43.21
& 60.74 & 55.91 \\

\rowcolor{blue!10}
\cellcolor{white}
& \textbf{\Buzzy\ (Ours)}
& \textbf{77.98} & \textbf{77.88}
& 61.40 & 61.58
& \textbf{59.47} & \textbf{41.67}
& \textbf{82.49} & \textbf{82.52}
& \textbf{46.60} & \textbf{46.29}
& \textbf{65.59} & \textbf{61.99} \\

\bottomrule
\end{tabular}%
}

% \caption{
% Main results across five multimodal MCQA benchmarks.
% We compare \Buzzy\ with greedy decoding, Prompt-enhanced Decoding,
% VCD, ICD, and ALC using five VLM backbones.
% We report accuracy (ACC) and macro-F1.
% Avg.\ denotes the unweighted mean across the five benchmarks.
% The best result for each metric within each backbone is highlighted in
% \textbf{bold}. 
% }
\caption{Main results across five multimodal MCQA benchmarks and five VLM backbones. We compare \Buzzy\ with greedy decoding, prompt-enhanced decoding, VCD, ICD, and ALC using accuracy (ACC) and Macro-F1. Avg.\ denotes the unweighted mean across benchmarks, and bold indicates the best result within each backbone.}
\label{tab:comprehensive_results}
\end{table*}

\section{Experiments}
\subsection{Experimental Setup}

\paragraph{Benchmarks and Backbone Models.}
We evaluate \Buzzy\ on five multimodal MCQA benchmarks: MAC~\citep{jiang2025mac}, SciFIBench~\citep{roberts2024scifibench}, MMSci~\citep{li2024mmsci}, MMBench~\citep{liu2024mmbench}, and MMStar~\citep{chen2024we}. These benchmarks cover scientific figures, general visual reasoning, and diverse multimodal question types. They are designed to require genuine visual grounding, with correct answers depending primarily on evidence from the visual input. We use five vision-language backbones: Qwen3.5-4B, Qwen3.5-9B~\citep{qwen3.5}, Phi-3.5-vision-instruct~\citep{abdin2024phi3technicalreporthighly}, Molmo-7B-D~\citep{deitke2024molmopixmoopenweights}, and GLM-4.6V-Flash~\citep{vteam2025glm45vglm41vthinkingversatilemultimodal}. The two Qwen models enable comparison across scales within the same family, while the remaining models represent different architectures. Detailed benchmark descriptions and evaluation settings are provided in Appendix~\ref{app:benchmarks}.

\paragraph{Baselines and Implementation Details.}
We compare \Buzzy\ with greedy decoding, prompt-enhanced decoding, VCD~\citep{leng2024vcd}, ICD~\citep{wang2024icd}, and Adaptive Logit Correction (ALC)~\citep{atabuzzaman2025benchmarkingmitigatingmcqaselection}. Greedy decoding uses the original multimodal input, while prompt-enhanced decoding emphasizes visual evidence through instruction. VCD and ICD construct contrastive branches using a degraded image and a disturbance instruction, respectively, whereas ALC targets structural choice bias by correcting estimated preferences for option labels or positions. Baseline configurations and prompts are detailed in Appendix~\ref{app:baselines}. 
% We report accuracy and Macro-F1, adopting the original settings for VCD, ICD, and ALC. 
% We evaluate all methods using accuracy and Macro-F1, while following the original implementation settings for VCD, ICD, and ALC. 
All experiments are conducted using an NVIDIA A100 GPU (80GB VRAM).  We fix $\alpha=0.5$ for \Buzzy\ without dataset-specific tuning.

\subsection{Results}

As shown in Table~\ref{tab:comprehensive_results}, \Buzzy\ achieves the highest average accuracy and Macro-F1 for all five backbones. Averaged across the benchmarks and backbones, it obtains 57.84 accuracy and 53.56 Macro-F1. This consistency across model families and scales suggests that the correction is not specific to a particular architecture. \Buzzy\ also leads on MMBench for every backbone. Since MMBench uses circular evaluation to control for positional bias, these gains extend beyond preferences for fixed answer labels. ALC remains below \Buzzy\ for every backbone, suggesting that correcting option-label or positional biases does not fully address semantic text priors. Prompt-enhanced decoding also yields inconsistent gains, indicating that visual-grounding instructions alone are insufficient.

Although VCD remains the strongest aggregate baseline, it consistently falls below \Buzzy\ in average performance. The largest improvement appears with GLM-4.6V-Flash, where \Buzzy\ increases average accuracy from 62.40 to 65.59 and Macro-F1 from 57.81 to 61.99. On MMSci, \Buzzy\ achieves the highest accuracy for every backbone and exceeds VCD by 6.55 points with GLM-4.6V-Flash. VCD and ICD also require an additional multimodal forward pass, while \Buzzy\ uses a lighter text-only reference branch. These results show that discounting text-only preferences improves MCQA performance without a second visual pass.

% \begin{table}[hbt!]
%   \centering
%   \footnotesize
%   \setlength{\tabcolsep}{4pt}
%   \renewcommand{\arraystretch}{1.08}
%   \begin{tabularx}{\columnwidth}{
%     @{}
%     >{\raggedright\arraybackslash}p{0.28\columnwidth}
%     >{\raggedright\arraybackslash}X
%     @{}
%   }
%     \toprule
%     \textbf{Diagnostic} & \textbf{Definition and interpretation} \\
%     \midrule

%     \textbf{Gold uplift} &
%     \makecell[l]{
%       $l_{\mathrm{buzzy}}(\gold)-l_{\mm}(\gold)$ \\[-1pt]
%       Gold-score change after correction
%     } \\
%     \midrule

%     \textbf{Visual margin} &
%     \makecell[l]{
%       $l_{\mm}(\gold)-l_{\txt}(\gold)$ \\[-1pt]
%       Multimodal vs.\ text-only gold gap
%     } \\
%     \midrule

%     \textbf{Text-prior hit} &
%     \makecell[l]{
%       $\mathbf{1}\!\left[\argmax_{c}p_{\txt}(c)=\gold\right]$ \\[-1pt]
%       Whether the text-only top-1 is gold
%     } \\
%     \midrule

%     \textbf{Prior alignment} &
%     \makecell[l]{
%       $\cos(p_{\mm},p_{\txt})$ \\[-1pt]
%       Multimodal and text-only similarity
%     } \\

%     \bottomrule
%   \end{tabularx}
%   \caption{Diagnostic quantities used in the analysis.}
%   \label{tab:diagnostics}
% \end{table}

\section{Analysis}
\subsection{Does BUZZY Expose Image-Conditioned Evidence?}
To mathematically analyze how subtracting the text-prior
in candidate-logit space promotes visual reasoning, we decompose the
multimodal candidate preference into text-aligned and orthogonal components. We analyze both \emph{corrected} and \emph{harmed} cases, where
\Buzzy\ respectively fixes or overturns the greedy prediction.

% \subsection{Text-Aligned and Text-Orthogonal Components}
% Let $\mathbf{m}$ and $\mathbf{t}$ denote the mean-centered
% multimodal and text-only candidate-logit vectors. Mean-centering removes the shared offset across candidates while
% preserving their relative logit differences. We regard
% $\mathbf{t}$ as the direction of the model's text-only candidate
% preference and project the multimodal preference $\mathbf{m}$ onto
% this direction:
% \begin{equation}
% \beta
% =
% \frac{\mathbf{m}^{\top}\mathbf{t}}
% {\mathbf{t}^{\top}\mathbf{t}},
% \qquad
% \mathbf{r}
% =
% \mathbf{m}-\beta\mathbf{t}.
% \end{equation}
% Here, $\beta$ is a scalar projection coefficient that measures how
% strongly the multimodal candidate preference follows the text-only
% direction. The vector $\beta\mathbf{t}$ is therefore the
% text-aligned component of the multimodal preference, while
% $\mathbf{r}$ is the remaining component after removing this
% projection. By construction,
% \begin{equation}
% m=\beta t+r,
% \qquad
% r^\top t=0,
% \end{equation}
% so $\mathbf{r}$ is orthogonal to the text-only preference direction. 
\begin{table*}[t!]
\centering
\footnotesize
\renewcommand{\arraystretch}{1.05} % 줄 간격 미세 축소 (기존 1.1)
\setlength{\tabcolsep}{3.5pt} % 열 간격 미세 축소 (기존 4.5pt)

\begin{adjustbox}{max width=0.98\textwidth} % 너비를 약간 더 타이트하게 제한
% 가운데 두 케이스 사이의 간격도 2em -> 1.5em으로 살짝 줄임
\begin{tabular}{@{} l ccc ccc @{\hspace{1.5em}} ccc ccc @{}}
\toprule
& 
\multicolumn{6}{c}{\textbf{Corrected Cases}} & 
\multicolumn{6}{c}{\textbf{Harmed Cases}} \\
\cmidrule(lr){2-7} \cmidrule(l){8-13}

& 
\multicolumn{3}{c}{GOD rate (\%)} & 
\multicolumn{3}{c}{Residual margin} & 
\multicolumn{3}{c}{GOD rate (\%)} & 
\multicolumn{3}{c}{Residual margin} \\
\cmidrule(lr){2-4} \cmidrule(lr){5-7} \cmidrule(lr){8-10} \cmidrule(l){11-13}

Dataset & 
Orig & Hard & Rand & Orig & Hard & Rand & 
Orig & Hard & Rand & Orig & Hard & Rand \\
\midrule

MAC 
& 70.6 & 11.8 & 35.3 & $+0.489$ & $-0.595$ & $-0.377$ 
& 21.4 & 0.0 & 7.1 & $-2.149$ & $-1.367$ & $-0.975$ \\

SciFIBench 
& 81.7 & 45.1 & 42.3 & $+0.532$ & $-0.130$ & $-0.153$ 
& 21.7 & 13.0 & 28.3 & $-0.542$ & $-0.702$ & $-0.476$ \\

MMSci 
& 83.9 & 62.1 & 55.9 & $+0.472$ & $+0.137$ & $+0.014$ 
& 17.8 & 16.2 & 25.9 & $-0.621$ & $-0.554$ & $-0.342$ \\

MMBench 
& 75.3 & 58.6 & 49.0 & $+0.460$ & $+0.198$ & $-0.046$ 
& 11.7 & 19.8 & 28.4 & $-0.481$ & $-0.388$ & $-0.236$ \\

MMStar 
& 85.2 & 61.1 & 58.3 & $+0.530$ & $+0.258$ & $+0.049$ 
& 24.0 & 22.7 & 29.3 & $-0.461$ & $-0.479$ & $-0.543$ \\
\midrule

\textbf{Overall} 
& \textbf{80.8} & \textbf{58.1} & \textbf{52.3} & $\mathbf{+0.483}$ & $\mathbf{+0.132}$ & $\mathbf{-0.025}$ 
& \textbf{17.2} & \textbf{17.6} & \textbf{27.0} & $\mathbf{-0.586}$ & $\mathbf{-0.524}$ & $\mathbf{-0.369}$ \\
\bottomrule
\end{tabular}
\end{adjustbox}

\caption{
\textbf{Visual Grounding Analysis of \Buzzy.}
The Gold-Over-Distractor (GOD) rate measures the proportion of
examples in which the text-orthogonal residual favors the gold
answer over the fixed competing candidate. A positive residual margin indicates stronger residual support for the gold
answer.
}
\label{tab:residual_image_replacement}
\end{table*}
\subsubsection{Text-Aligned and Orthogonal Components}
Let $m$ and $t$ denote the mean-centered candidate logit vectors
from the multimodal and text-only branches.
% Mean-centering removes the common offset across candidates while
% preserving their relative preferences.
Geometrically, we treat $t$ as an axis representing the model's
text-only candidate preference. We then project $m$ onto this axis
to measure how much of the multimodal preference follows the text-only
candidate pattern.  The remaining component $r$ is orthogonal to $t$, capturing variation in the multimodal preference that is uncorrelated with the text-only candidate pattern. Formally, we decompose $m$ into a component aligned with $t$ and
an orthogonal residual $r$:
\begin{equation}
\beta
=
\frac{m^{\top}t}{t^{\top}t},
\qquad
r
=
m-\beta t.
\end{equation}
Here, $\beta$ determines the magnitude of the projection, and
$\beta t$ is the part of the multimodal preference that follows
the text-only candidate pattern. The residual $r$ captures the non-textual component, whose visual
sensitivity we examine next. These components satisfy
\begin{equation}
m=\beta t+r,
\qquad
r^{\top}t=0.
\end{equation}
A larger $\beta$ indicates stronger alignment with the text-only direction, while \Buzzy\ reduces the component by subtracting $\alpha t$:
$m-\alpha t=(\beta-\alpha)t+r$.

\subsubsection{Visual Grounding Analysis of \Buzzy}
\label{sec:visual-grounding-analysis}
% We test whether the text-orthogonal residual preserved by \Buzzy\ reflects image-grounded evidence.
% We compare three conditions:
% (i) the \emph{original} image matched to the question;
% (ii) a \emph{random} image sampled from another example in the same benchmark; and
% (iii) a \emph{hard-negative} image retrieved as the most similar image in the same benchmark using CLIP
% cosine similarity. 
% The question, answer choices, and text-only direction remain unchanged in all conditions. We measure its preference between the
% gold answer and the fixed greedy distractor using
% \begin{equation}
% M_{\mathrm{res},i}(x)
% =
% r_{i,y_i^*}(x)-r_{i,d_i}(x).
% \end{equation}
% We also report the Gold-Over-Distractor (GOD) rate, defined as
% the proportion of corrected examples with a positive residual
% margin:
% \begin{equation}
% R_{\mathrm{GOD}}(x)
% =
% \frac{1}{|\mathcal{S}_{\mathrm{corr}}|}
% \sum_{i\in\mathcal{S}_{\mathrm{corr}}}
% \mathbb{I}
% \left[
% M_{\mathrm{res},i}(x)>0
% \right].
% \end{equation}

We test whether the text-orthogonal residual contains image-specific
evidence by replacing the visual input. We compare three conditions:
the original image, a hard negative retrieved as the most similar
non-matching image within the same benchmark using CLIP cosine
similarity, and a random image from the same benchmark.
For each example, we compare the gold answer $y_i^*$ with a fixed
competing answer $d_i$, defined as the greedy error in corrected cases
and the \Buzzy\ error in harmed cases. This answer is selected under
the original image and held fixed across conditions. We compute the
residual margin
\begin{equation}
M_{\mathrm{res},i}(x)
=
r_{i,y_i^*}(x)-r_{i,d_i}(x),
\end{equation}
where a positive value favors the gold answer, and report its positive
rate for each group $g\in\{\mathrm{corr},\mathrm{harm}\}$:
\begin{equation}
R_{\mathrm{GOD}}^{g}(x)
=
\frac{1}{|\mathcal{S}_{g}|}
\sum_{i\in\mathcal{S}_{g}}
\mathbb{I}\!\left[M_{\mathrm{res},i}(x)>0\right].
\end{equation}

% As shown in Table~\ref{tab:residual_image_replacement}, corrected
% cases exhibit an aggregate Original $>$ Hard Negative $>$ Random
% pattern: the GOD rate decreases from 80.8\% to 58.1\% and 52.3\%,
% while the residual margin decreases from $+0.483$ to $+0.132$ and
% $-0.025$. Harmed cases show the complementary trend: the GOD rate
% increases from 17.2\% to 17.6\% and 27.0\%, while the residual
% margin moves toward zero from $0.586$ to $0.524$ and $0.369$.
% As the matched image is
% progressively replaced by a hard negative and then a random image,
% the residual margin moves toward zero in both corrected and harmed
% cases. This graded attenuation indicates that the residual captures
% image-specific candidate evidence. By suppressing the text-aligned
% component, \Buzzy\ gives this visually grounded signal greater
% relative influence on the final prediction.

Table~\ref{tab:residual_image_replacement} shows an aggregate
Original $>$ Hard Negative $>$ Random pattern in corrected cases.
Across both corrected and harmed cases, image replacement weakens
the residual preference and moves its margin toward zero. Thus, the residual preference progressively weakens toward zero as
image relevance decreases from the matched image to a hard negative
and then to a random image, indicating that $r$ captures
image-conditioned evidence.
Overall, results show that \Buzzy\ shifts  prediction away
from text-aligned preferences and towards image-conditioned residual evidence.

\begin{figure}[t]
    \centering

    % =========================
    % (a) Theoretical cost
    % =========================
    \begin{minipage}[t]{0.47\linewidth} % [t]로 상단 기준 정렬
        \centering
        \vspace{0pt} % minipage 최상단 기준점 고정
        \small 
        \renewcommand{\arraystretch}{1.8}
        \setlength{\tabcolsep}{3pt}

        \begin{adjustbox}{max width=\linewidth}
        \begin{tabular}{l c c}
            \toprule
            \textbf{Method}
            & \textbf{Prefill Passes}
            & \textbf{Theoretical Cost} \\
            \midrule

            Greedy
            & $1\times$ MM
            & $\mathcal{O}((L_q+L_v)^2d)$ \\

            VCD
            & $2\times$ MM
            & $\mathcal{O}(2(L_q+L_v)^2d)$ \\

            ICD
            & $2\times$ MM
            & $\mathcal{O}(2(L_q+L_v)^2d)$ \\

            \rowcolor{blue!10}
            \textbf{\method{}}
            & $\mathbf{1\times MM + 1\times Text}$
            & $\mathbf{\mathcal{O}((L_q+L_v)^2d+L_q^2d)}$ \\

            \bottomrule
        \end{tabular}
        \end{adjustbox}

        % 표가 짧아서 생기는 빈 공간을 채워 (a) 라벨을 오른쪽 (b) 라벨과 높이 맞춤
        \vspace{20pt} 
        {\small\textbf{(a)} Theoretical prefill cost}
    \end{minipage}
    \hfill
    % =========================
    % (b) Latency plot
    % =========================
    \begin{minipage}[t]{0.49\linewidth} % [t]로 상단 기준 정렬
        \centering
        \vspace{0pt} % minipage 최상단 기준점 고정 (오른쪽 그래프의 윗선과 왼쪽 표 윗선이 완벽히 일치함)

        \begin{tikzpicture}
            \begin{axis}[
                width=\linewidth,
                height=4.2cm,
                xlabel={Latency (s)},
                ylabel={Avg. Accuracy (\%)},
                xmin=0.18, xmax=0.52,
                ymin=51, ymax=59,
                grid=major,
                grid style={dashed, gray!30},
                label style={font=\scriptsize},
                tick label style={font=\scriptsize},
                clip=false
            ]

            \tikzstyle{labelstyle}=[
                font=\scriptsize,
                fill=white,
                inner sep=1pt,
                fill opacity=0.85,
                text opacity=1
            ]

            \node[circle, fill=green!80, inner sep=1.7pt]
                (greedy) at (axis cs:0.220949,54.452) {};
            \node[above left=3pt, xshift=8pt, labelstyle] at (greedy) {Greedy};

            \node[circle, fill=gray!75!black, inner sep=1.7pt]
                (prompt) at (axis cs:0.220949,53.984) {};
            \node[below=4pt, xshift=8pt, labelstyle] at (prompt) {Prompt-enhanced};

            \node[circle, fill=orange!85!black, inner sep=1.7pt]
                (alc) at (axis cs:0.244847,54.212) {};
            \node[right=3pt, labelstyle] at (alc) {ALC};

            \node[
                star,
                star points=5,
                star point ratio=2.25,
                fill=red!80,
                inner sep=3pt
            ] (ours) at (axis cs:0.296002,57.836) {};

            \node[
                right=11pt,
                font=\scriptsize\bfseries,
                fill=white,
                inner sep=1pt,
                fill opacity=0.85,
                text opacity=1
            ] at (ours) {Buzzy};

            \node[circle, fill=purple!80, inner sep=1.7pt]
                (icd) at (axis cs:0.416445,51.564) {};
            \node[above=2pt, labelstyle] at (icd) {ICD};

            \node[circle, fill=blue!80, inner sep=1.7pt]
                (vcd) at (axis cs:0.450466,55.848) {};
            \node[above=2pt, labelstyle] at (vcd) {VCD};

            \end{axis}
        \end{tikzpicture}

        \vspace{4pt}
        {\small\textbf{(b)} Accuracy--latency trade-off}
    \end{minipage}

    \vspace{8pt}
    \caption{
    Efficiency analysis of \Buzzy.
    (a) Theoretical prefill cost, where MM and Text denote multimodal
    and text-only passes, respectively.
    (b) Average accuracy--latency trade-off across evaluated models.
    \Buzzy\ avoids a second multimodal pass while achieving the highest
    average accuracy.
    }
    \label{fig:efficiency_analysis}
    \vspace{-13pt}
\end{figure}
\subsection{Efficiency Analysis}
\label{sec:efficiency}

Figure~\ref{fig:efficiency_analysis} compares the theoretical prefill costs of the decoding methods. We use $L_q$ for the number of textual tokens in the question, choices, and prompt, $L_v$ for the number of visual tokens, and $d$ for the hidden dimension. Under the standard self-attention cost model, one multimodal prefill requires $O((L_q+L_v)^2d)$. VCD and ICD construct their reference predictions using an additional multimodal condition, resulting in two full multimodal passes and a total cost of $O(2(L_q+L_v)^2d)$. 
% Because visual inputs often contain many tokens, repeating this computation introduces substantial overhead. 
In contrast, \Buzzy\ combines one multimodal pass with one text-only pass at a total cost of $O((L_q+L_v)^2d+L_q^2d)$, avoiding additional computation over the visual tokens.

Figure~\ref{fig:efficiency_analysis} shows that this theoretical advantage translates to empirical latency on average across all evaluated models. VCD and ICD approximately double the latency of greedy decoding from $0.22$\,s to $0.45$\,s and $0.42$\,s, respectively. ALC has a low per-example latency of $0.24$\,s but lower average accuracy than greedy decoding. This latency also excludes the upfront calibration passes used to estimate the selection-bias vector. In comparison, \Buzzy\ reaches 57.84\% average accuracy with a latency of $0.30$\,s, improving greedy decoding by 3.38 points with only $0.08$\,s of additional latency. These results demonstrate that \Buzzy\ provides the optimal balance between performance and efficiency; by running the multimodal forward pass only once, it avoids severe latency overheads while successfully preserving the accuracy gains of contrastive decoding to achieve the highest overall performance.

% Domain-specific fine-tuning table moved to appendix for ICLR 2027 page-limit compliance.

\subsection{Sensitivity to \texorpdfstring{$\alpha$}{Alpha}}
\label{sec:alpha_sensitivity}

We analyze the sensitivity of \Buzzy\ to the subtraction weight
$\alpha$ across all five backbones and five benchmarks.

% Insert the wrapped figure after the paragraph has started,
% rather than immediately after the subsection heading.
\begin{wrapfigure}{r}{0.44\linewidth}
    \vspace{-6pt}
    \centering

    % Local spacing for the wrapped figure
    \setlength{\intextsep}{2pt}
    \setlength{\columnsep}{8pt}

    \begin{tikzpicture}
        \begin{axis}[
            width=\linewidth,
            height=4.00cm,
            xlabel={Subtraction weight $\alpha$},
            ylabel={Average score (\%)},
            xmin=0.05,
            xmax=0.95,
            ymin=50,
            ymax=59,
            xtick={0.1,0.3,0.5,0.7,0.9},
            ytick={50,52,54,56,58},
            grid=major,
            grid style={dashed, gray!30},
            label style={font=\scriptsize},
            tick label style={font=\tiny},
            legend style={
                at={(0.5,-0.37)},
                anchor=north,
                legend columns=-1,
                draw=none,
                font=\tiny,
                /tikz/every even column/.append style={
                    column sep=3pt
                }
            }
        ]

        % Average accuracy
        \addplot+[
            line width=1.2pt,
            mark=*,
            mark size=2pt
        ] coordinates {
            (0.1,55.58)
            (0.3,57.01)
            (0.5,57.84)
            (0.7,57.32)
            (0.9,55.57)
        };

        % Average Macro-F1
        \addplot+[
            line width=1.2pt,
            mark=*,
            mark size=2pt
        ] coordinates {
            (0.1,51.50)
            (0.3,52.84)
            (0.5,53.56)
            (0.7,52.63)
            (0.9,50.81)
        };

        % Default alpha = 0.5
        \addplot[
            dotted,
            thick,
            forget plot
        ] coordinates {
            (0.5,50)
            (0.5,59)
        };

        \legend{Accuracy, Macro-F1}

        \end{axis}
    \end{tikzpicture}

    \caption{
    Sensitivity of \Buzzy\ to $\alpha$.
    Moderate values remain stable, with $\alpha=0.5$
    achieving the best aggregate performance.
    }
    \label{fig:alpha_sensitivity_all}

    \vspace{-5pt}
\end{wrapfigure}

Here, $\alpha$ controls the strength of the text-prior penalty
subtracted from the multimodal branch. We initially set
$\alpha=0.5$ to match the weighting ratio used by the default
configurations of VCD and ICD, ensuring a fair comparison under
the same relative weighting. We further examine the robustness of
this choice across $\alpha \in \{0.1, 0.3, 0.5, 0.7, 0.9\}$.

As shown in Figure~\ref{fig:alpha_sensitivity_all}, moderate values
from $0.3$ to $0.7$ yield comparable aggregate performance.
In particular, $\alpha=0.5$ achieves the highest overall accuracy
and Macro-F1, with scores of 57.84 and 53.56, respectively.
A smaller weight of $0.1$ provides limited correction, whereas
the aggressive setting of $0.9$ reduces accuracy and Macro-F1
to 55.57\% and 50.81\%.

This pattern reflects the trade-off controlled by $\alpha$:
insufficient subtraction leaves misleading textual preferences
largely unchanged, whereas excessive subtraction can suppress
useful textual support. We therefore use $\alpha=0.5$ throughout
the main experiments as a fixed global setting, without
backbone- or dataset-specific tuning.

This variation suggests that VLMs retain text-only preferences to
different degrees and respond differently to stronger correction. We
therefore use $\alpha=0.5$ throughout the main experiments as a fixed
global setting, avoiding backbone- or dataset-specific tuning.
Notably, the setting chosen to match VCD and ICD also achieves the best
aggregate performance, providing empirical support for the consistency
of this shared weighting ratio. This consistency supports our choice of
$\alpha$ and enables a fair comparison with these baselines.

\subsection{Generalizability to Domain-Specific Fine-Tuned Models}
\label{sec:finetuned_models_main}

To examine whether the gains of \Buzzy\ remain after targeted model adaptation, we evaluate on Qwen3-VL-8B-Instruct-ChartQA-sft, a version of Qwen3-VL-8B further fine-tuned with LoRA on 7,398 human-annotated ChartQA-style examples for three epochs, updating 21.8M parameters. This setting provides a stronger science- and chart-aligned backbone and therefore tests whether decoding-time text-prior correction remains useful even when visual reasoning has already been improved through domain-specific fine-tuning.

\begin{table*}[t]
\centering
\small
\renewcommand{\arraystretch}{1.10}
\setlength{\tabcolsep}{4pt}
\resizebox{\textwidth}{!}{%
\begin{tabular}{l cc cc cc cc cc cc}
\toprule
\multirow{2}{*}{\textbf{Method}}
& \multicolumn{2}{c}{\textbf{MAC}}
& \multicolumn{2}{c}{\textbf{SciFIBench}}
& \multicolumn{2}{c}{\textbf{MMSci}}
& \multicolumn{2}{c}{\textbf{MMBench}}
& \multicolumn{2}{c}{\textbf{MMStar}}
& \multicolumn{2}{c}{\textbf{Avg.}} \\
\cmidrule(lr){2-3}
\cmidrule(lr){4-5}
\cmidrule(lr){6-7}
\cmidrule(lr){8-9}
\cmidrule(lr){10-11}
\cmidrule(lr){12-13}
& ACC & F1
& ACC & F1
& ACC & F1
& ACC & F1
& ACC & F1
& ACC & F1 \\
\midrule
Greedy
& 80.73 & 80.83
& 66.90 & 67.48
& 63.30 & 49.66
& 83.95 & 83.97
& 48.33 & 47.46
& 68.64 & 65.88 \\

Prompt-enhanced
& 78.29 & 78.18
& 66.20 & 66.13
& 63.73 & 50.45
& 83.60 & 83.62
& 48.07 & 47.22
& 67.98 & 65.12 \\

VCD
& 81.96 & 82.10
& 66.80 & 67.37
& 65.13 & 50.03
& 84.04 & 84.05
& 49.20 & 48.59
& 69.43 & 66.43 \\

ICD
& 80.73 & 80.83
& 67.00 & 66.87
& 62.33 & 49.22
& 83.18 & 83.19
& 47.47 & 46.88
& 68.14 & 65.40 \\

ALC
& 75.54 & 75.48
& 65.10 & 65.01
& 62.30 & 47.48
& 84.01 & 84.01
& 47.60 & 47.01
& 66.91 & 63.80 \\

\rowcolor{blue!10}
\textbf{\Buzzy\,(Ours)}
& \textbf{82.26} & \textbf{82.50}
& \textbf{67.50} & \textbf{67.96}
& \textbf{67.72} & \textbf{52.37}
& \textbf{84.57} & \textbf{84.57}
& \textbf{50.60} & \textbf{50.15}
& \textbf{70.53} & \textbf{67.51} \\
\bottomrule
\end{tabular}%
}
\caption{Performance comparison on Qwen3-VL-8B-Instruct-ChartQA-sft. \Buzzy\ consistently outperforms the baselines across all five benchmarks. Avg. denotes the unweighted mean across benchmarks.}
\label{tab:qwen3_chartqa_main}
\end{table*}

As shown in Table~\ref{tab:qwen3_chartqa_main}, \Buzzy\ achieves the best accuracy and Macro-F1 on all five benchmarks. It reaches 70.53 average accuracy and 67.51 Macro-F1, outperforming the strongest baseline, VCD, by 1.10 and 1.08 points, respectively. These gains indicate that candidate-induced semantic text priors can persist even after domain-specific fine-tuning, and that fine-tuning and decoding-time prior correction provide complementary benefits rather than redundant ones.

\subsection{Statistical Comparison}
\label{sec:statistic}

To test whether the aggregate gains reflect consistent example-level
improvements, we compare \Buzzy\ with each baseline using McNemar's
test~\citep{mcnemar1947note}. This paired test examines examples on
which two methods disagree and determines whether \Buzzy\ corrects
more baseline errors than it introduces.

\begin{wraptable}{r}{0.45\linewidth}
    \vspace{-6pt}
    \centering
    \setlength{\intextsep}{3pt}
    \setlength{\columnsep}{8pt}

    \scriptsize
    \renewcommand{\arraystretch}{1.3}
    \setlength{\tabcolsep}{3.2pt}

    \begin{adjustbox}{max width=\linewidth}
    \begin{tabular}{@{}lccc@{}}
        \toprule
        \textbf{Dataset}
        & \textbf{vs Greedy}
        & \textbf{vs VCD}
        & \textbf{vs ICD} \\
        \midrule
        MAC
        & \textbf{0.04877}
        & \textbf{0.01135}
        & \textbf{3.02e-09} \\

        SciFIBench
        & \textbf{0.02608}
        & \textbf{0.007656}
        & \textbf{1.08e-08} \\

        MMSci
        & \textbf{4.51e-14}
        & \textbf{4.75e-05}
        & \textbf{1.12e-37} \\

        MMBench
        & \textbf{0.000141}
        & 0.07499
        & \textbf{2.79e-23} \\

        MMStar
        & \textbf{0.02204}
        & 0.1542
        & \textbf{3.68e-06} \\
        \bottomrule
    \end{tabular}
    \end{adjustbox}

    \caption{
    McNemar's test $p$-values on Qwen3.5-4B.
    Bold indicates $p<0.05$.
    }
    \label{tab:p_values_qwen4b}

    \vspace{-5pt}
\end{wraptable}

Table~\ref{tab:p_values_qwen4b} reports the resulting $p$-values
across the five benchmarks. \Buzzy\ shows significant improvements
over greedy decoding and ICD on all five benchmarks. Against VCD,
the difference is significant on MAC, SciFIBench, and MMSci, but not
on MMBench or MMStar.

Combined with the higher accuracy of \Buzzy, these results indicate
that its corrections exceed its newly introduced errors in most
comparisons. The nonsignificant results do not establish equivalence
with VCD, but indicate weaker evidence where the observed gains are
smaller. Overall, the analysis supports a consistent advantage over
greedy decoding and ICD and a more dataset-dependent advantage over
VCD.

\section{Conclusion}

% In this work, we introduced \Buzzy, a training-free decoding method for multimodal MCQA that discounts text-only candidate preferences from multimodal scores. Across five VLM backbones and five benchmarks, \Buzzy\ achieved the highest average accuracy and Macro-F1 for every backbone. Our analysis showed that corrections are concentrated in cases where the text-only branch favors an incorrect candidate while the multimodal branch retains support for the gold answer. \Buzzy\ also reduces inference latency by over 41\% compared with contrastive methods that require a second multimodal pass. These findings support our central view that visual grounding is better characterized by candidate support relative to text-only preference than by absolute multimodal confidence, establishing the text-only context as an effective and efficient reference for multimodal decoding.

In this work, we introduced \Buzzy, a training-free decoding method for multimodal MCQA that discounts text-only candidate preferences from multimodal scores. Across five VLMs and five benchmarks, \Buzzy\ achieved the highest average accuracy and Macro-F1 for every backbone while reducing latency by over 28\% relative to multimodal contrastive methods. Our analysis showed that gains concentrate when misleading text-only preferences obscure multimodal support for the gold answer. These results support measuring visual grounding through multimodal support relative to text-only preference rather than absolute confidence. More broadly, our findings establish the text-only context as a simple yet effective control signal for identifying choice-induced bias and supporting more reliable, visually grounded multimodal decision making.
\section*{Limitations}
\label{sec:limitations}
% While \Buzzy\ demonstrates robust performance improvements across various benchmarks, it has a few limitations. First, \Buzzy\ is most effective when visual evidence provides a strong signal that diverges from the text prior. However, when visual evidence is ambiguous, subtracting the text prior can be counterproductive—especially if the prior already points to the correct answer. 

% Second, our method requires direct access to the model's pre-softmax logits to compute the contrastive penalty during the decoding phase. Consequently, \Buzzy\ cannot be directly applied to proprietary, closed-source models (e.g., GPT-5, Claude Fable 5, or Gemini 3) accessed via standard APIs, as these black-box systems typically do not expose token-level probability distributions or allow custom interventions in the decoding process.

While \Buzzy\ provides an effective post-hoc correction, it operates
only on the final candidate logits after the full forward pass.
Therefore, it cannot reveal how text-induced bias emerges or
interacts with visual evidence across intermediate layers and
attention patterns. Future work could extend this analysis with
layer-wise attribution and internal representation analysis.

\Buzzy\ also requires access to candidate-level pre-softmax logits, or equivalent token-level scores, under both multimodal and text-only contexts. It therefore cannot be directly applied through black-box APIs that expose only generated outputs or insufficient token-level probabilities. Its applicability is consequently limited by the scoring interface provided by the model, rather than by whether the model itself is publicly available.

\clearpage
\section*{AI Use Statement}
Generative AI tools were used during the implementation process to assist in identifying potential bugs in the code and during the preparation of this manuscript for grammar and language correction. The tools were not used to generate research results, conduct analyses, or make scientific conclusions.

\bibliographystyle{iclr2027_conference}
\bibliography{references}

\clearpage
\appendix

\section{Experimental Details}
\subsection{Benchmark Details}
\label{app:benchmarks}

We evaluate our method on five multimodal MCQA benchmarks spanning scientific figure understanding and general visual perception and reasoning. These tasks are designed to require genuine visual grounding, such that correct predictions rely primarily on evidence contained in the image.

\paragraph{MAC}
MAC~\citep{jiang2025mac} is a live benchmark constructed from cover images and associated descriptions published in leading scientific journals. Rather than focusing on literal chart reading, it requires models to connect abstract visual elements in journal covers with high-level scientific narratives. Its periodically updated snapshots are designed to reflect evolving scientific topics and model capabilities.

\paragraph{SciFIBench}
SciFIBench~\citep{roberts2024scifibench} evaluates scientific figure interpretation using figure--caption pairs collected from arXiv papers. It includes two complementary tasks that require selecting the correct caption for a figure or the correct figure for a caption. Hard negatives are selected through adversarial filtering and subsequently verified by humans, making superficial figure--text matching less effective.

\paragraph{MMSci}
MMSci~\citep{li2024mmsci} is constructed from peer-reviewed articles in \textit{Nature Communications} and covers diverse scientific figures beyond conventional charts, including microscopy images, schematics, experimental visualizations, and molecular structures. Its multiple-choice benchmark evaluates whole-figure caption matching, subfigure-to-subcaption matching, and the reverse matching of subcaptions to subfigures. These tasks require both holistic figure understanding and fine-grained localization of scientific content.

\paragraph{MMBench}
MMBench~\citep{liu2024mmbench} is a general-purpose benchmark covering visual perception, attribute and relation reasoning, and logical reasoning. Its hierarchical capability taxonomy enables evaluation across both broad and fine-grained visual skills. MMBench also introduces CircularEval, which cyclically changes the order of answer choices and requires consistent predictions across permutations to reduce positional effects.

\paragraph{MMStar}
MMStar~\citep{chen2024we} is designed to address visual redundancy and data leakage in existing multimodal benchmarks. Candidate questions are filtered using language and vision-language models, followed by human review to retain samples that require visual information and span different difficulty levels. It therefore provides a purified setting for evaluating whether performance gains arise from multimodal understanding rather than text-only solvability or memorization.

\subsection{Jensen--Shannon Divergence}
\label{app:jsd}

To quantify the discrepancy between answer preferences under different input conditions, we compute the Jensen--Shannon (JS) divergence~\citep{lin1991divergence} between the multimodal and text-only answer distributions. Specifically, let
$P=p_{\mathrm{mm}}(\cdot\mid x,q,\mathcal{C})$ and
$Q=p_{\mathrm{txt}}(\cdot\mid q,\mathcal{C})$ denote the two distributions over the same candidate set. Their JS divergence is defined as
\begin{equation}
\begin{aligned}
M
&=
\frac{P+Q}{2}, \\[2pt]
\mathrm{JS}(P\|Q)
&=
\frac{1}{2}\mathrm{KL}(P\|M)
+
\frac{1}{2}\mathrm{KL}(Q\|M).
\end{aligned}
\end{equation}
where $\mathrm{KL}(\cdot\|\cdot)$ denotes the Kullback--Leibler divergence,
\begin{equation}
\mathrm{KL}(P\|M)
=
\sum_i P(i)\log\frac{P(i)}{M(i)}.
\end{equation}

JS divergence is symmetric and bounded. Lower values indicate closer alignment between the multimodal and text-only preferences, whereas higher values indicate a greater discrepancy between them. We use this discrepancy to examine how the multimodal prediction departs from the preference induced by the textual context alone.

\subsection{Additional JS Divergence Distributions}
\label{sec:additional_js}

\begin{figure}[t]
    \centering
    \includegraphics[
        width=0.72\linewidth
    ]{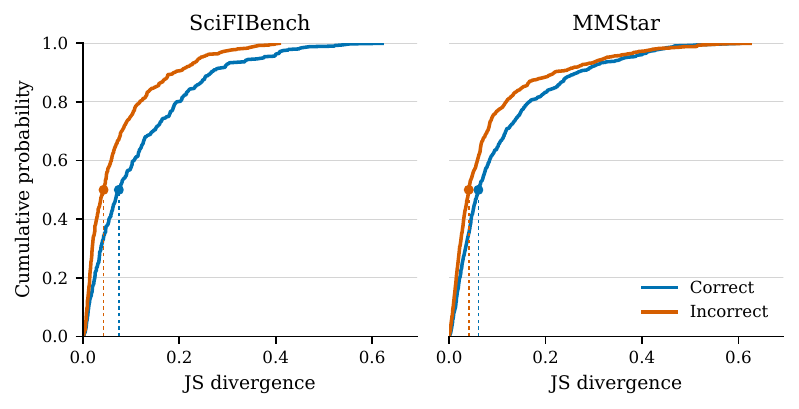}

    \caption{
    Cumulative distributions of JS divergence for correct and incorrect
    predictions on SciFIBench and MMStar. Both benchmarks show the same
    overall trend as the three benchmarks reported in
    Figure~\ref{fig:motivation_js_ecdf}, with correct predictions shifted
    toward higher divergence from the text-only distribution.
    }
    \label{fig:motivation_js_ecdf_remaining}
\end{figure}

Figure~\ref{fig:motivation_js_ecdf_remaining} reports the
JS-divergence distributions for SciFIBench and MMStar, complementing
the three representative benchmarks shown in
Figure~\ref{fig:motivation_js_ecdf}. Both benchmarks exhibit the same
overall pattern: correct predictions are shifted toward higher
JS divergence than incorrect predictions, with higher median
divergence at a cumulative probability of $0.5$. These results further
support that correct multimodal predictions tend to depart more
strongly from text-only preferences across all five benchmarks.

\subsection{Hyperparameter Details}
\label{app:hyperparameter}

We use the default hyperparameter settings from the released implementations of VCD and ICD. Both methods perform contrastive decoding at the candidate-logit level and can be expressed using the common scoring rule
\begin{equation}
s_{\mathrm{CD}}(c)
=
(1+\lambda)l_{\mathrm{orig}}(c)
-
\lambda l_{\mathrm{ref}}(c),
\end{equation}
where $l_{\mathrm{ref}}(c)$ is obtained from a noise-corrupted image for VCD and a disturbed instruction for ICD. We use the default $\lambda=1.0$, resulting in an original-to-reference weighting ratio of $2\!:\!-1$.

\Buzzy\ instead uses the text-only branch as its reference and computes
\begin{equation}
s_{\mathrm{Buzzy}}(c)
=
l_{\mathrm{mm}}(c)
-
\alpha l_{\mathrm{txt}}(c).
\end{equation}
We set $\alpha=0.5$ throughout the main experiments without dataset-specific tuning. Multiplying these scores by two does not change their ranking, making the default form rank-equivalent to $2l_{\mathrm{mm}}(c)-l_{\mathrm{txt}}(c)$. Thus, \Buzzy\ uses the same effective $2\!:\!-1$ weighting ratio as the default VCD and ICD configurations while replacing their constructed reference inputs with the text-only context.

\begin{table*}[t]
\centering
\small
\setlength{\tabcolsep}{5pt}
\renewcommand{\arraystretch}{1.08}
\resizebox{\textwidth}{!}{%
\begin{tabular}{llccccc}
\toprule
\textbf{Backbone}
& \textbf{Dataset}
& \textbf{vs. Greedy}
& \textbf{vs. VCD}
& \textbf{vs. ICD}
& \textbf{vs. ALC}
& \textbf{vs. Prompt-Enhanced} \\
\midrule

\multirow{5}{*}{Qwen3.5-4B}
& MAC
& \textbf{0.04877}
& \textbf{0.01135}
& $\mathbf{3.02\times10^{-9}}$
& \textbf{0.04296}
& $\mathbf{4.37\times10^{-5}}$
\\

& SciFIBench
& \textbf{0.02608}
& \textbf{0.007656}
& $\mathbf{1.08\times10^{-8}}$
& 0.4044
& $\mathbf{1.91\times10^{-7}}$
\\

& MMSci
& $\mathbf{4.51\times10^{-14}}$
& $\mathbf{4.75\times10^{-5}}$
& $\mathbf{1.12\times10^{-37}}$
& $\mathbf{4.25\times10^{-8}}$
& $\mathbf{2.39\times10^{-28}}$
\\

& MMBench
& \textbf{0.000141}
& 0.07499
& $\mathbf{2.79\times10^{-23}}$
& \textbf{0.001897}
& $\mathbf{1.76\times10^{-26}}$
\\

& MMStar
& \textbf{0.02204}
& 0.1542
& $\mathbf{3.68\times10^{-6}}$
& 0.1258
& \textbf{0.001258}
\\

\midrule

\multirow{5}{*}{Qwen3.5-9B}
& MAC
& 0.4531
& 1
& 0.3877
& \textbf{0.02266}
& 0.7905
\\

& SciFIBench
& \textbf{0.003462}
& 0.05044
& $\mathbf{5.73\times10^{-6}}$
& $\mathbf{7.33\times10^{-5}}$
& \textbf{0.001888}
\\

& MMSci
& $\mathbf{2.87\times10^{-27}}$
& $\mathbf{1.03\times10^{-7}}$
& $\mathbf{5.46\times10^{-18}}$
& $\mathbf{5.08\times10^{-20}}$
& $\mathbf{2.12\times10^{-27}}$
\\

& MMBench
& $\mathbf{2.48\times10^{-12}}$
& \textbf{0.008085}
& $\mathbf{1.06\times10^{-35}}$
& $\mathbf{8.30\times10^{-11}}$
& \textbf{0.000572}
\\

& MMStar
& \textbf{0.000477}
& \textbf{0.000585}
& $\mathbf{2.81\times10^{-5}}$
& \textbf{0.000221}
& \textbf{0.001016}
\\

\midrule

\multirow{5}{*}{Phi-3.5}
& MAC
& \textbf{0.003838}
& \textbf{0.007943}
& \textbf{0.007149}
& \textbf{0.03648}
& 0.07039
\\

& SciFIBench
& $\mathbf{4.16\times10^{-10}}$
& 0.1249
& $\mathbf{2.66\times10^{-13}}$
& $\mathbf{1.01\times10^{-10}}$
& $\mathbf{1.34\times10^{-8}}$
\\

& MMSci
& $\mathbf{2.96\times10^{-13}}$
& 0.1939
& $\mathbf{4.68\times10^{-22}}$
& $\mathbf{1.83\times10^{-14}}$
& $\mathbf{1.35\times10^{-10}}$
\\

& MMBench
& \textbf{0.01609}
& 0.2245
& $\mathbf{7.62\times10^{-8}}$
& \textbf{0.002392}
& \textbf{0.001393}
\\

& MMStar
& \textbf{0.002155}
& 0.7861
& \textbf{0.002052}
& \textbf{0.01552}
& 0.3352
\\

\midrule

\multirow{5}{*}{Molmo-7B-D}
& MAC
& \textbf{0.000472}
& 0.1325
& \textbf{0.000717}
& \textbf{0.000606}
& 0.5327
\\

& SciFIBench
& 0.5264
& 0.1486
& \textbf{0.000215}
& 0.8138
& 0.313
\\

& MMSci
& \textbf{0.00182}
& 0.6167
& $\mathbf{1.98\times10^{-10}}$
& \textbf{0.000715}
& 0.3169
\\

& MMBench
& $\mathbf{2.55\times10^{-10}}$
& 0.3466
& $\mathbf{6.17\times10^{-43}}$
& $\mathbf{3.60\times10^{-5}}$
& 0.826
\\

& MMStar
& \textbf{0.01224}
& 0.7985
& \textbf{0.000118}
& 0.05831
& 0.8202
\\

\midrule

\multirow{5}{*}{GLM-4.6V-Flash}
& MAC
& \textbf{0.006611}
& \textbf{0.000313}
& $\mathbf{1.96\times10^{-6}}$
& $\mathbf{1.29\times10^{-5}}$
& \textbf{0.00018}
\\

& SciFIBench
& 0.3634
& 0.6992
& $\mathbf{3.93\times10^{-9}}$
& 0.3112
& 0.1238
\\

& MMSci
& $\mathbf{9.46\times10^{-37}}$
& $\mathbf{2.08\times10^{-19}}$
& $\mathbf{1.69\times10^{-66}}$
& $\mathbf{6.59\times10^{-46}}$
& $\mathbf{1.58\times10^{-40}}$
\\

& MMBench
& $\mathbf{3.18\times10^{-11}}$
& $\mathbf{1.13\times10^{-5}}$
& $\mathbf{2.21\times10^{-20}}$
& $\mathbf{3.64\times10^{-11}}$
& $\mathbf{7.54\times10^{-27}}$
\\

& MMStar
& 0.1325
& 0.2759
& \textbf{0.02418}
& 0.09938
& \textbf{0.000499}
\\

\bottomrule
\end{tabular}%
}

\caption{
McNemar's test $p$-values comparing \Buzzy\ with
baseline decoding methods using paired item-level predictions.
Bold values indicate statistical significance at $p<0.05$ before
multiple-comparison correction.
}
\label{tab:significance_results_raw}
\label{tab:significance_results}
\end{table*}

\section{Baseline Details}
\label{app:baselines}

% We compare \Buzzy with five training-free baselines.
% All methods use the same backbone, evaluation samples, answer options,
% and decoding configuration.
% Appendix~\ref{app:prompts}.

We compare \Buzzy\ with five training-free baselines. All methods use the same backbone, evaluation samples, answer options, and candidate-label scoring procedure. Method-specific prompts are provided in Appendix~\ref{app:prompts}.

\paragraph{Greedy Decoding.}
% Greedy decoding scores the candidate answer options using the original
% multimodal prompt and image, and selects the option with the highest
% next-token log-probability. We use deterministic decoding with
% temperature set to 0. The method does not alter the output
% distribution or use an auxiliary branch for prediction.

Greedy decoding scores the candidate labels using the original multimodal input and selects the candidate with the highest next-token logit. We use deterministic argmax decoding without sampling. This baseline neither modifies the output distribution nor introduces an auxiliary branch.

\paragraph{Prompt-Enhanced Decoding.}
% This baseline adds an explicit visual-grounding instruction to the
% original prompt and performs a single multimodal forward pass.
% It evaluates whether prompt-level guidance alone can reduce reliance on
% textual priors.

Prompt-enhanced decoding adds an explicit visual-grounding instruction to the base prompt and performs a single multimodal forward pass. It tests whether instruction-level guidance alone can reduce reliance on textual preferences. The complete instruction is provided in Appendix~\ref{prompt:prompt-enhanced}.

\paragraph{VCD}
% Visual Contrastive Decoding (VCD)~\citep{leng2024vcd} contrasts the
% candidate-option distributions obtained from the original image and a
% visually degraded counterpart, while keeping the textual prompt identical
% in both branches. In our implementation, the degraded image is generated
% by adding diffusion-style Gaussian noise to the RGB input image. We first
% normalize the image to $[0,1]$, sample $\epsilon \sim \mathcal{N}(0,I)$,
% and construct
% $\tilde{x}_t=\sqrt{\bar{\alpha}_t}x+\sqrt{1-\bar{\alpha}_t}\epsilon$
% using a 1000-step sigmoid beta schedule. We set the noise step to
% $t=500$ for all experiments and clamp the resulting image back to
% $[0,1]$ before inference.

VCD contrasts candidate logits obtained from the original image and a visually degraded counterpart while keeping the textual prompt fixed. We generate the degraded image by normalizing the RGB input to $[0,1]$, sampling $\epsilon\sim\mathcal{N}(0,I)$, and applying
\begin{equation}
\tilde{x}_t
=
\sqrt{\bar{\alpha}_t}x
+
\sqrt{1-\bar{\alpha}_t}\epsilon
\end{equation}
under a 1,000-step sigmoid beta schedule. We use $t=500$ throughout and clamp the resulting image to $[0,1]$ before inference.

\paragraph{ICD}
% Instruction Contrastive Decoding (ICD)~\citep{wang2024icd} contrasts
% the candidate-option distribution from the original prompt with a
% distribution obtained under a hallucination-inducing instruction.
% The visual input is kept unchanged in both branches, while the disturbed
% branch modifies only the textual prompt by adding an instruction that
% encourages a fuzzy or confused interpretation of the image. ICD then
% compares the option logits from the original and disturbed prompts to
% suppress answers that remain highly likely under the hallucination-
% inducing condition.

ICD contrasts candidate logits from the original prompt with those obtained under a hallucination-inducing instruction. Both branches use the same image, while the disturbed branch adds an instruction encouraging a fuzzy or confused interpretation of the visual content. ICD penalizes candidates that remain strongly preferred under this disturbed condition. The disturbed instruction is provided in Appendix~\ref{app:prompts}.

\paragraph{ALC}
% Adaptive Logit Correction (ALC)~\citep{atabuzzaman2025benchmarkingmitigatingmcqaselection}
% estimates a selection-bias vector over answer options and subtracts it
% from the original multimodal logits before prediction.
% The correction weight is chosen adaptively based on the confidence of
% the original output distribution.

ALC estimates an ensemble selection-bias vector using general and contextual calibration prompts. It subtracts this vector from the original multimodal logits, with the correction strength determined by the confidence of the original prediction. Unlike \Buzzy, ALC targets structural preferences for particular option labels or positions rather than semantic preferences induced by candidate content.

\section{Prompt Details}
\label{app:prompts}

All methods use the same base prompt for the target multimodal question,
except for Prompt-Enhanced Decoding and the disturbed branch of ICD.
Specifically, Greedy, VCD, \Buzzy, ALC, and the original branch of ICD
share the prompt below. VCD modifies only the visual input, while the
text-only branch of \Buzzy\ uses the same textual template with the image
removed. ALC also applies the base prompt to the target question, but
uses an additional semantically empty prompt to estimate selection bias.

\begin{samepage}
\begin{promptbox}{Base Prompt}
Question: \{question\}

\medskip
Choose exactly one option (\{A/B/C/...\}) that best matches the image.

\medskip
A. \{option A\}

B. \{option B\}

...

\medskip
Answer with one letter only: A, B, C, ...
\end{promptbox}

\paragraph{Prompt-Enhanced Decoding}
Prompt-Enhanced Decoding modifies the base prompt by adding an explicit
instruction that encourages the model to prioritize visual evidence over
textual plausibility or option wording. All other components of the
question and option template remain unchanged.

\begin{promptbox}{Prompt-Enhanced Prompt}
Question: \{question\}

\medskip
Choose exactly one option (\{A/B/C/...\}) that best matches the image.

\medskip
Use the image as the primary evidence when choosing among the options.
Do not rely on textual plausibility or option wording as bias when it
conflicts with the image.

\medskip
A. \{option A\}

B. \{option B\}

...

\medskip
Answer with one letter only: A, B, C, ...
\label{prompt:prompt-enhanced}
\end{promptbox}
\end{samepage}

\paragraph{ICD}
% The original branch of ICD uses the common base prompt. Its disturbed
% branch prepends a hallucination-inducing instruction to the same
% image-question and option template, while leaving the remaining prompt
% structure unchanged.

The original ICD branch uses the base prompt. For the disturbed branch,
we prepend the N1 disturbance instruction from the original implementation
while leaving the image, question, and answer choices unchanged.

\begin{promptbox}{ICD Disturbed-Branch Prompt}
For visual understanding, follow this additional instruction:

\medskip
I want you avoid any specific identification or categorization of the
objects depicted.

\medskip
Question: \{question\}

\medskip
Choose exactly one option (\{A/B/C/...\}) that best matches the image.

\medskip
A. \{option A\}

B. \{option B\}

...

\medskip
Answer with one letter only: A, B, C, ...
\end{promptbox}

\paragraph{ALC}
Following the original implementation, ALC estimates both general and contextual selection biases. For each candidate count $K$, we construct semantically empty calibration prompts with the same $K$ option labels as the target examples. The general bias is estimated from these prompts, whereas the contextual bias is estimated from a randomly selected subset of benchmark examples without using their gold labels. The two mean-centered bias vectors are averaged and adaptively subtracted from the target logits according to prediction confidence.

\begin{promptbox}{ALC Calibration Prompt}
Question: Select one answer option. The options contain no task-specific
information.

\medskip
Choose exactly one option (A/B/C/D).

\medskip
A. Option

B. Option

C. Option

D. Option

\medskip
Answer with one letter only: A, B, C, D.
\end{promptbox}

\section{Statistical Details}
\label{app:statistics}

% We conduct pairwise McNemar's tests~\citep{mcnemar1947note} between \Buzzy\ and each baseline using paired item-level predictions on the same evaluation examples. McNemar's test compares the number of discordant cases, namely, examples corrected by \Buzzy\ and examples answered correctly by the baseline but changed incorrectly by \Buzzy. A $p$-value below $0.05$ indicates a statistically significant difference under the null hypothesis that the two methods have the same error rate.

We conduct pairwise McNemar's tests between \Buzzy\ and each baseline using predictions on the same evaluation examples. The test evaluates whether the two types of discordant cases occur equally often, namely, errors corrected by \Buzzy\ and correct baseline predictions overturned by \Buzzy. We report $p$-values and use $p<0.05$ as the nominal significance threshold under the null hypothesis that the methods have equal error rates.

% Overall, the accuracy gains of \Buzzy\ are statistically significant in most backbone--benchmark comparisons. \Buzzy\ significantly outperforms Greedy in 21 of 25 settings and ICD in 24 of 25 settings, indicating that its improvements are generally consistent at the item level. Comparisons with VCD are more mixed, as the two methods often produce similar predictions and relatively small accuracy differences; nevertheless, \Buzzy\ achieves significant gains in 10 settings. Importantly, the only cases in the main results where \Buzzy\ obtains lower accuracy than VCD---Molmo-7B-D and GLM-4.6V-Flash on SciFIBench---are not statistically significant ($p=0.1486$ and $p=0.6992$, respectively). More generally, most non-significant comparisons correspond to small absolute performance differences, suggesting no reliable degradation relative to the corresponding baseline.

At the nominal $0.05$ level, \Buzzy\ significantly outperforms greedy decoding in 21 of 25 backbone--benchmark settings and ICD in 24 of 25 settings. Together with its higher accuracy, these results indicate that \Buzzy\ generally corrects more errors than it introduces. The evidence against VCD is less uniform, with significant improvements in 10 settings. \Buzzy\ has lower accuracy than VCD only on SciFIBench with Molmo-7B-D and GLM-4.6V-Flash, and neither difference is significant, with $p=0.1486$ and $p=0.6992$, respectively. Thus, the tests provide broad evidence of improvement over greedy decoding and ICD, while the advantage over VCD remains dependent on the backbone and benchmark.

\section{Case Study: Gold-Answer Recovery}
\label{appendix:case_study}

We provide qualitative analyses of gold-answer recovery across the five benchmarks. In each case, greedy decoding follows a distractor favored by the text-only branch, whereas \Buzzy\ suppresses this preference and recovers the gold answer. The examples complement our aggregate analysis by showing how text-only preferences affect individual predictions.

\subsection{Gold-Answer Recovery in MAC}
\begin{figure*}[t]
    \centering
    % 왼쪽 이미지 영역
    \begin{minipage}[c]{0.45\textwidth}
        \centering
        \includegraphics[width=0.6\linewidth]{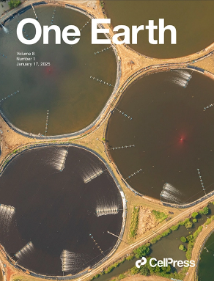}
    \end{minipage}% <-- 주석 기호(%)를 달아 의도치 않은 띄어쓰기를 방지합니다.
    \hspace{0.02\textwidth} % 간격 고정
    % 오른쪽 그래프 영역
    \begin{minipage}[c]{0.42\textwidth}
    \centering
    \begin{tikzpicture}
        \begin{axis}[
            width=\linewidth,
            height=4.0cm,         % <-- 그래프의 세로 높이 고정
            ybar=0.5pt,
            bar width=6pt,        % <-- 막대 두께
            enlarge x limits={abs=12pt},
            ylabel={Probability ($p$)},
            symbolic x coords={A, B, C, D},
            xtick=data,
            point meta=y,
            nodes near coords,
            every node near coord/.append style={
                rotate=90, 
                anchor=west, 
                font=\fontsize{5}{6}\selectfont, % 라벨 폰트
                /pgf/number format/fixed,
                /pgf/number format/precision=2
            },
            ymax=1.2,
            legend style={
                at={(0.5,-0.25)}, 
                anchor=north, 
                legend columns=-1, 
                font=\fontsize{6}{7}\selectfont, 
                draw=none
            },
            title={\footnotesize Comparison of Answer Distributions},
            label style={font=\scriptsize},
            tick label style={font=\scriptsize}
        ]
            % Multimodal Branch
            \addplot coordinates {(A,0.6063) (B,0.3677) (C,0.0162) (D,0.0098)};
            % Text-only Prior
            \addplot coordinates {(A,0.8531) (B,0.0057) (C,0.0794) (D,0.0618)};
            % Buzzy
            \addplot coordinates {(A,0.1171) (B,0.8656) (C,0.0102) (D,0.0070)};
            
            \legend{Multimodal, Text-only, \method{}}
        \end{axis}
    \end{tikzpicture}
    \end{minipage}

    \vspace{0.2em} % <-- 위아래 간격
    
    % 하단 텍스트 박스
    \begin{minipage}[c]{0.92\textwidth} 
    \footnotesize
    \begin{tcolorbox}[colback=white, colframe=gray!20, arc=0mm, boxrule=0.5pt, left=2pt, right=2pt, top=1pt, bottom=1pt]
        \textbf{Question:} Which of the following options best describe the cover image? \quad
        \textbf{Correct Answer: \color{blue}B}
        \begin{itemize}[leftmargin=1.2em, itemsep=0pt, parsep=0pt, topsep=2pt] 
            \item \textbf{Option A:} California is home to expansive, water-intensive industrial agriculture. As the state faced severe drought in the 2010s, industry remained buoyed by unjust water rights, while neighboring families suffered without access to water. The cover by Tali Weinberg transforms lush agricultural landscapes into drought-stricken fields.
            \item \textbf{Option B (Gold):} Mangroves are crucial carbon sinks, yet these ecosystems are being lost rapidly, mainly due to aquaculture and shrimp farming. The cover highlights the restoration potential of lost mangroves in China and Southeast Asia.
            \item \textbf{Option C:} Clean water is vital to planetary health, yet anthropogenic pressures threaten this resource. The cover shows the swirls of an algae bloom caused by agricultural runoff, with consequences for biodiversity and human health.
            \item \textbf{Option D:} The inaugural issue of \textit{One Earth} focuses on climate action, depicting the generation at risk from climate change and their engagement with the issue.
        \end{itemize}
    \end{tcolorbox}
    \end{minipage}
    \caption{A MAC example in which \method{} suppresses the text-prior-dominant distractor selected by greedy decoding and recovers the gold answer.}
    \label{fig:case_study_mac}
\end{figure*}

Figure~\ref{fig:case_study_mac} shows a MAC example in which the cover
depicts large circular aquaculture ponds surrounded by green coastal
vegetation, supporting Option B's description of mangrove loss caused
by shrimp farming and the potential for ecosystem restoration. However,
the multimodal branch selects Option A, a plausible environmental
narrative about drought-stricken agriculture. The text-only branch
favors A even more strongly, while assigning almost no probability to
the gold answer B, indicating that the incorrect prediction is largely
reinforced by textual preference. In contrast, B gains substantial
support only when the image is provided. \Buzzy\ discounts the
text-prior-dominant score of A and recovers the visually grounded gold
answer B.

% 1. MAC Example
\subsection{Gold-Answer Recovery in SciFIBench}
\begin{figure*}[t]
    \centering
    % 왼쪽 이미지 영역
    \begin{minipage}[c]{0.45\textwidth}
        \centering
        \includegraphics[width=0.9\linewidth]{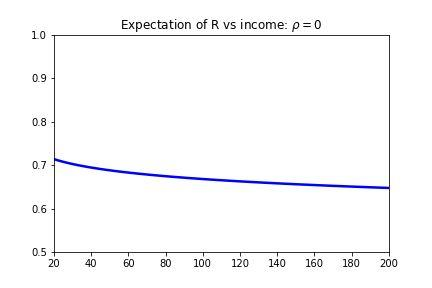}
        \vspace{0.2em}
        {\scriptsize Scientific figure example\par}
    \end{minipage}%
    \hspace{0.02\textwidth}
    % 오른쪽 그래프 영역
    \begin{minipage}[c]{0.42\textwidth}
    \centering
    \begin{tikzpicture}
        \begin{axis}[
            width=\linewidth,
            height=4.0cm,
            ybar=0.5pt,
            bar width=5.5pt,      % 5개 항목이므로 두께를 아주 약간만 줄여 안정감 확보
            enlarge x limits={abs=12pt},
            ylabel={Probability ($p$)},
            symbolic x coords={A, B, C, D, E},
            xtick=data,
            point meta=y,
            nodes near coords,
            every node near coord/.append style={
                rotate=90, 
                anchor=west, 
                font=\fontsize{5}{6}\selectfont,
                /pgf/number format/fixed,
                /pgf/number format/precision=2
            },
            ymax=1.2, 
            legend style={
                at={(0.5,-0.25)}, 
                anchor=north, 
                legend columns=-1, 
                font=\fontsize{6}{7}\selectfont, 
                draw=none
            },
            title={\footnotesize Comparison of Answer Distributions},
            label style={font=\scriptsize},
            tick label style={font=\scriptsize}
        ]
            % Multimodal Branch
            \addplot coordinates {(A,0.404) (B,0.029) (C,0.519) (D,0.043) (E,0.004)};
            % Text-only Prior
            \addplot coordinates {(A,0.004) (B,0.022) (C,0.918) (D,0.040) (E,0.017)};
            % Buzzy (Contrastive)
            \addplot coordinates {(A,0.870) (B,0.026) (C,0.071) (D,0.028) (E,0.005)};
            
            \legend{Multimodal, Text-only, \method{}}
        \end{axis}
    \end{tikzpicture}
    \end{minipage}

    \vspace{0.2em}
    
    % 하단 텍스트 박스
    \begin{minipage}[c]{0.92\textwidth} 
    \footnotesize
    \begin{tcolorbox}[colback=white, colframe=gray!20, arc=0mm, boxrule=0.5pt, left=2pt, right=2pt, top=1pt, bottom=1pt]
        \textbf{Question:} Which caption best matches the image? \quad
        \textbf{Correct Answer: \color{blue}A}
        \begin{itemize}[leftmargin=1.2em, itemsep=0pt, parsep=0pt, topsep=2pt]
            \item \textbf{Option A (Gold):} $\rho=0$ case. Left panel depicts the conditional expectation function $\mathbb{E}\{\mathbb{E}[R_i \mid X_{1i}=y, X_{2i}]\}$ as a function of $y$, when $\rho=0$. The observable relationship between $R$ and income reflects the true negative effect $\beta_1$. The right panel reports regression results on a simulated dataset of 10,000 observations.
            \item \textbf{Option B:} $\rho=1$ case. The regression results reflect a spurious positive association between income and reported satisfaction, and Column (3) reports an infeasible direct regression recovering the true parameters.
            \item \textbf{Option C:} Mean response $R_i$ versus income $X_{1i}$, comparing $\rho=0$ and $\rho=1$; the right panel reflects a spurious upward slope under correlated reporting heterogeneity.
            \item \textbf{Option D:} $\rho=0$ case with a lowess regression of $R$ on income in a simulated dataset of 10,000 observations.
            \item \textbf{Option E:} Radial case with $d=2$ and $m=1$, analyzing $\alpha_\star^{\mathrm{rad},(n=0)}(p)$ and $\Lambda_D^{\mathrm{rad},(n=0)}(\alpha,p)$.
        \end{itemize}
    \end{tcolorbox}
    \end{minipage}
    \caption{A SciFIBench example in which \method{} suppresses a scientifically plausible distractor favored by the text-only branch and recovers the gold answer.}
    \label{fig:case_study_scifi}
\end{figure*}

Figure~\ref{fig:case_study_scifi} presents a SciFIBench example whose
scientific figure is labeled as the $\rho=0$ setting and shows a gradually
decreasing relationship between the expected response and income. This
visual pattern supports Option A, which identifies the $\rho=0$ case and
describes the negative effect of income on the response. Nevertheless,
greedy decoding selects Option C: the multimodal branch assigns
probabilities of 0.52 and 0.40 to C and A, respectively. Although C is a
scientifically plausible description, it refers to a comparison between
the $\rho=0$ and $\rho=1$ settings and a spurious upward relationship,
which are not supported by the displayed figure. The text-only branch
favors C even more strongly ($p=0.92$), while assigning almost no
probability to the gold answer A. This indicates that C is preferred
primarily because of its plausible scientific terminology rather than
the visual evidence. By discounting this text-only preference, \Buzzy\
suppresses C and promotes the visually supported gold answer A
($p=0.87$).
% 2. SciFIBench Example

\subsection{Gold-Answer Recovery in MMSci}
\begin{figure*}[t]
    \centering
    % 1. 왼쪽 이미지 영역
    \begin{minipage}[c]{0.45\textwidth}
        \centering
        \includegraphics[width=0.9\linewidth]{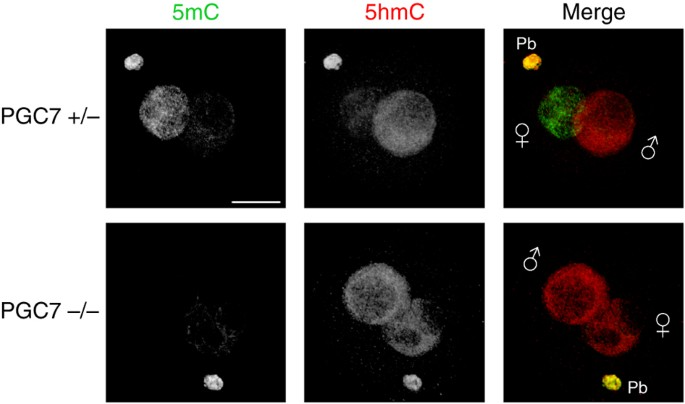}
    \end{minipage}% <-- 주석 기호(%)를 달아 의도치 않은 띄어쓰기를 방지합니다.
    % 2. 널찍한 \hfill 대신 고정된 좁은 간격(\hspace)을 사용합니다.
    \hspace{0.02\textwidth} % 간격을 더 좁히려면 0.01\textwidth, 넓히려면 0.05\textwidth 등으로 조절하세요.
    % 3. 오른쪽 그래프 영역
    \begin{minipage}[c]{0.42\textwidth} % 너비를 0.4에서 0.42로 살짝 늘려 안정감을 줍니다.
    \centering
    \begin{tikzpicture}
        \begin{axis}[
            width=\linewidth,
            height=4.0cm,         % <-- 그래프의 세로 높이 고정
            ybar=0.5pt,
            bar width=6pt,        % <-- 막대 두께
            enlarge x limits={abs=12pt},
            ylabel={Probability ($p$)},
            symbolic x coords={A, B, C, D},
            xtick=data,
            point meta=y,
            nodes near coords,
            every node near coord/.append style={
                rotate=90, 
                anchor=west, 
                font=\fontsize{5}{6}\selectfont, % 라벨 폰트
                /pgf/number format/fixed,
                /pgf/number format/precision=2
            },
            ymax=1.2,
            legend style={
                at={(0.5,-0.25)}, 
                anchor=north, 
                legend columns=-1, 
                font=\fontsize{6}{7}\selectfont, 
                draw=none
            },
            title={\footnotesize Comparison of Answer Distributions},
            label style={font=\scriptsize},
            tick label style={font=\scriptsize}
        ]
            % Multimodal Branch
            \addplot coordinates {(A,0.013) (B,0.235) (C,0.724) (D,0.028)};
            % Text-only Prior
            \addplot coordinates {(A,0.015) (B,0.036) (C,0.914) (D,0.036)};
            % Buzzy
            \addplot coordinates {(A,0.048) (B,0.552) (C,0.335) (D,0.066)};
            
            \legend{Multimodal, Text-only, \Buzzy}
        \end{axis}
    \end{tikzpicture}
    \end{minipage}

    \vspace{0.2em} % <-- 위아래 간격
    
    % 하단 텍스트 박스
    \begin{minipage}[c]{0.92\textwidth} 
    \footnotesize
    \begin{tcolorbox}[colback=white, colframe=gray!20, arc=0mm, boxrule=0.5pt, left=2pt, right=2pt, top=1pt, bottom=1pt]
        \textbf{Question:} Which of the following captions best describes the whole figure? \quad
        \textbf{Correct Answer: \color{blue}B}
        \begin{itemize}[leftmargin=1.2em, itemsep=0pt, parsep=0pt, topsep=2pt] 
            \item \textbf{Option A:} Relative expression of Tet transcripts in development.
            \item \textbf{Option B (Gold):} Increase of 5hmC in the maternal genome of zygotes derived from PGC7-null oocytes.
            \item \textbf{Option C:} 5hmC preferentially appears in the paternal genome of early mouse preimplantation embryos.
            \item \textbf{Option D:} 5hmC is present in rabbit and bovine zygotes.
        \end{itemize}
    \end{tcolorbox}
    \end{minipage}
    \caption{An MMSci example in which \method{} suppresses a distractor favored by both branches and recovers the gold answer.}
    \label{fig:case_study_mmsci}
\end{figure*}

Figure~\ref{fig:case_study_mmsci} presents an MMSci example comparing
5mC and 5hmC signals in zygotes derived from PGC7 wild-type and
PGC7-null oocytes. In the lower PGC7-null row, the figure shows an
increased 5hmC signal in the maternal genome, supporting Option B.
Nevertheless, greedy decoding selects Option C: the multimodal branch
assigns probabilities of 0.72 and 0.24 to C and B, respectively.
Option C is scientifically plausible because it describes the known
asymmetric distribution of 5hmC between the parental genomes, but it
does not capture the genotype-specific comparison emphasized in the
figure. The text-only branch favors C even more strongly
($p=0.91$), while assigning little probability to the gold answer B
($p=0.04$), indicating that the distractor is strongly reinforced by
its textual plausibility. By discounting this shared text-only
preference, \Buzzy\ reduces the probability of C and promotes the
visually supported gold answer B from 0.24 to 0.55.

\subsection{Gold-Answer Recovery in MMBench}
Figure~\ref{fig:case_study_mmbench_action} presents a more discriminative MMBench recovery example with four descriptive answer choices. The image shows a taekwondo practice scene, but the text-only branch strongly favors a plausible yet visually unsupported fishing description. Greedy decoding follows this prior and predicts Option A. In contrast, \method{} discounts the question-only preference and shifts probability mass to Option D, the gold answer, with a large margin over the nearest distractor. This example illustrates that prior subtraction can produce a clear recovery even when the answer options are long, semantically rich descriptions rather than short object labels.

% 3. MMBench Action Example
\begin{figure*}[t]
    \centering
    % 왼쪽 이미지 영역
    \begin{minipage}[c]{0.45\textwidth}
        \centering
        \includegraphics[width=0.9\linewidth]{figure/mmbench_3001935_action.png}
    \end{minipage}%
    \hspace{0.02\textwidth}
    % 오른쪽 그래프 영역
    \begin{minipage}[c]{0.42\textwidth}
    \centering
    \begin{tikzpicture}
        \begin{axis}[
            width=\linewidth,
            height=4.0cm,
            ybar=0.5pt,
            bar width=6pt,
            enlarge x limits={abs=12pt},
            ylabel={Probability ($p$)},
            symbolic x coords={A, B, C, D},
            xtick=data,
            point meta=y,
            nodes near coords,
            every node near coord/.append style={
                rotate=90,
                anchor=west,
                font=\fontsize{5}{6}\selectfont,
                /pgf/number format/fixed,
                /pgf/number format/precision=2
            },
            ymax=1.3,
            ytick={0,0.2,0.4,0.6,0.8,1.0},
            legend style={
                at={(0.5,-0.25)}, 
                anchor=north, 
                legend columns=-1, 
                font=\fontsize{6}{7}\selectfont, 
                draw=none
            },
            title={\footnotesize Comparison of Answer Distributions},
            label style={font=\scriptsize},
            tick label style={font=\scriptsize}
        ]
            % Multimodal Branch
            \addplot coordinates {(A,0.8128) (B,0.0459) (C,0.0000) (D,0.1413)};
            % Text-only Prior
            \addplot coordinates {(A,0.9918) (B,0.0076) (C,0.0006) (D,0.0000)};
            % Buzzy
            \addplot coordinates {(A,0.0075) (B,0.0049) (C,0.0000) (D,0.9876)};
            
            \legend{Multimodal, Text-only, \method{}}
        \end{axis}
    \end{tikzpicture}
    \end{minipage}

    \vspace{0.2em}
    
    % 하단 텍스트 박스
    \begin{minipage}[c]{0.92\textwidth} 
    \footnotesize
    \begin{tcolorbox}[colback=white, colframe=gray!20, arc=0mm, boxrule=0.5pt, left=2pt, right=2pt, top=1pt, bottom=1pt]
        \textbf{Question:} What kind of human behavior does this picture describe? \quad
        \textbf{Correct Answer: \textcolor{blue}{D}}
        \begin{itemize}[leftmargin=1.2em, itemsep=0pt, parsep=0pt, topsep=2pt]
            \item \textbf{Option A:} A man is fishing on a riverbank, casting his line and waiting patiently for a bite while enjoying the tranquility of nature.
            \item \textbf{Option B:} A group of hikers are trekking through a wilderness trail, hiking over rugged terrain and encountering breathtaking vistas along the way.
            \item \textbf{Option C:} A man is doing calisthenics in his home gym, using his bodyweight to build strength and endurance through push-ups, squats, and other exercises.
            \item \textbf{Option D (Gold):} The man dressed in black was surrounded in the middle by ten men dressed in white who practiced taekwondo.
        \end{itemize}
    \end{tcolorbox}
    \end{minipage}
    \caption{An MMBench action-recognition case. Greedy decoding selects a text-prior-dominant distractor (Option A): the text-only branch assigns nearly all probability mass to a generic outdoor activity description, and the multimodal branch remains biased toward the same incorrect answer. After subtracting the text-only prior, \method{} suppresses A and assigns the highest probability to the visually grounded action description (Option D).}
    \label{fig:case_study_mmbench_action}
\end{figure*}

\subsection{Gold-Answer Recovery in MMStar}
\begin{figure*}[t]
    \centering
    \begin{minipage}[c]{0.48\textwidth}
        \centering
        \includegraphics[width=\linewidth]{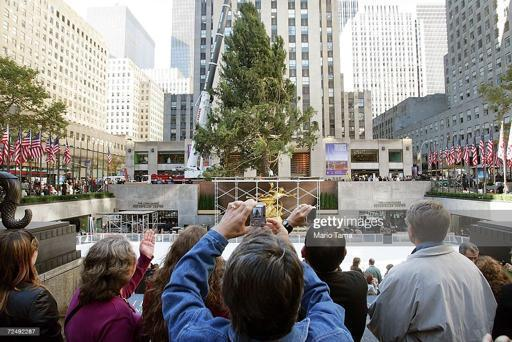}
    \end{minipage}
    \hspace{0.02\textwidth}
    % 오른쪽 그래프 영역
    \begin{minipage}[c]{0.42\textwidth}
    \centering
    \begin{tikzpicture}
        \begin{axis}[
            width=\linewidth,
            height=4.0cm,
            ybar=0.5pt,
            bar width=6pt,
            enlarge x limits={abs=12pt},
            ylabel={Probability ($p$)},
            symbolic x coords={A, B, C, D},
            xtick=data,
            point meta=y,
            nodes near coords,
            every node near coord/.append style={
                rotate=90,
                anchor=west,
                font=\fontsize{5}{6}\selectfont,
                /pgf/number format/fixed,
                /pgf/number format/precision=2
            },
            ymax=1.3,
            ytick={0,0.2,0.4,0.6,0.8,1.0},
            legend style={
                at={(0.5,-0.25)}, 
                anchor=north, 
                legend columns=-1, 
                font=\fontsize{6}{7}\selectfont, 
                draw=none
            },
            title={\footnotesize Comparison of Answer Distributions},
            label style={font=\scriptsize},
            tick label style={font=\scriptsize}
        ]
                % Multimodal Branch
                \addplot coordinates {(A,0.8774) (B,0.0110) (C,0.1115) (D,0.0000)};
                % Text-only Prior
                \addplot coordinates {(A,0.9982) (B,0.0012) (C,0.0006) (D,0.0000)};
                % Buzzy
                \addplot coordinates {(A,0.1507) (B,0.0555) (C,0.7898) (D,0.0040)};
                \legend{Multimodal, Text-only, \method{}}
            \end{axis}
        \end{tikzpicture}
    \end{minipage}

    \vspace{0.2em}

    \begin{small}
    \centering
    \begin{tcolorbox}[colback=white, colframe=gray!20, arc=0mm, boxrule=0.5pt, left=2pt, right=2pt, top=1pt, bottom=1pt]
        \raggedright
        \textbf{Question:} What is the primary scene depicted in the image?  \quad
        \textbf{Correct Answer: \textcolor{blue}{C}}
        \begin{itemize}[leftmargin=1.5em, itemsep=3pt, parsep=0pt]
            \item \textbf{Option A:} A group of people taking pictures of a statue in a plaza
            \item \textbf{Option B:} A large crowd taking pictures of each other in an outdoor event
            \item \textbf{Option C (Gold):} A group of people taking pictures of a Christmas tree in a plaza
            \item \textbf{Option D:} A group of people taking pictures of a famous building in New York City
        \end{itemize}
    \end{tcolorbox}
    \end{small}

    \caption{An MMStar example in which \method{} corrects a strong answer-option prior toward the wrong long caption, Option A, and recovers the visually grounded caption, Option C.}
    \label{fig:case_study_mmstar}
\end{figure*}

Figure~\ref{fig:case_study_mmstar} presents an MMStar example with long, sentence-level answer options. The image contains people photographing a large Christmas tree in a plaza, but both the multimodal and text-only branches strongly favor Option A, a plausible caption that shares several words with the correct answer. After correcting this answer-option prior, \method{} sharply suppresses the distractor and assigns the highest probability to the image-supported gold answer, Option C.

\section{Detailed Results on Benchmarks}
\label{appendix:scifibench_details}

To examine whether the aggregate gains of \Buzzy\ are driven by a small number of favorable categories, we report category-wise accuracy on SciFIBench, MMSci, MMBench, and MMStar. Across all four benchmarks, \Buzzy\ achieves the highest sample-weighted average accuracy with the same Qwen3.5-9B backbone. More importantly, its gains extend across scientific disciplines and visual reasoning capabilities with substantially different terminology, visual structures, and answer formats. These results indicate that the effectiveness of \Buzzy\ is not confined to a particular domain or question type, but instead reflects a broadly applicable benefit of discounting candidate preferences that can be explained by text alone.

On SciFIBench, \Buzzy\ achieves an overall accuracy of 58.00\%, outperforming the strongest baseline, VCD, by 2.10 percentage points. The improvement is observed in both major subsets: \Buzzy\ reaches 61.20\% on General Science, compared with 59.20\% for the strongest baseline, and 54.80\% on Computer Science, compared with 54.00\% for VCD. At the individual category level, \Buzzy\ obtains the best or tied-best performance in 12 of the 19 sub-categories. In General Science, it performs particularly strongly in economics, quantitative finance, statistics, mathematics, and physics, demonstrating that its advantage persists across both quantitative and conceptually diverse scientific fields. In Computer Science, \Buzzy\ leads the two largest categories, \textit{cross-list} and \textit{other cs}, while also obtaining the best or tied-best results in distributed computing, robotics, software engineering, computational linguistics, and artificial intelligence. The strong performance in the high-sample categories is especially important, as it shows that the aggregate improvement is supported by broad coverage rather than isolated gains in small subsets.

A similarly consistent pattern appears on MMSci. As shown in Table~\ref{tab:mmsci_category_results}, \Buzzy\ achieves the highest accuracy in four of the five scientific categories and improves the overall accuracy from 49.29\% with VCD to 52.14\%. The gains span biological sciences, physical sciences, health sciences, and earth and environmental sciences, with improvements over VCD of 2.23, 3.56, 2.73, and 4.06 percentage points, respectively. Since MMSci contains heterogeneous scientific visualizations, including microscopy images, experimental diagrams, molecular structures, and multi-panel figures, this consistency provides evidence that subtracting the text-only preference remains effective across substantially different forms of scientific visual evidence.

The results on the general-purpose benchmarks further demonstrate that the advantage of \Buzzy\ extends beyond scientific figures. On MMBench, \Buzzy\ achieves the highest overall accuracy of 79.10\%, exceeding VCD by 1.00 percentage point, and obtains the best or tied-best result in 11 of the 20 categories. Its strongest results include image-scene understanding, object localization, functional reasoning, action recognition, identity reasoning, and social-relation reasoning. These improvements cover both perceptual and reasoning-oriented capabilities, suggesting that \Buzzy\ does not merely improve recognition of visual content but also helps models use that content when distinguishing among semantically plausible answer choices.

On MMStar, \Buzzy\ improves the average accuracy from 41.00\% under greedy decoding to 43.70\%, yielding a 2.70-point gain over the strongest baseline. It achieves the best or tied-best performance in five of the six coarse categories, including coarse perception, fine-grained perception, instance reasoning, logical reasoning, and science and technology. The finer-grained results in Table~\ref{tab:qwen35_9b_mmstar_l2_category_breakdown} show particularly clear advantages in image emotion, image style and quality, recognition, localization, cross-instance attribute reasoning, single-instance reasoning, and diagram reasoning. These categories require different combinations of visual discrimination and semantic reasoning, further supporting the robustness of the proposed correction.

Overall, the category-wise results show that the superiority of \Buzzy\ is not attributable to a single benchmark, scientific field, or visual capability. Although individual baselines remain competitive in several specialized categories, none matches the consistently strongest aggregate performance of \Buzzy\ across all four benchmarks. This broad advantage supports our central hypothesis: candidate choices introduce textual preferences that persist across domains, and explicitly discounting these preferences allows the model to place greater weight on evidence uniquely contributed by the image.

\begin{table*}[t]
\centering
% 1. 폰트 크기 축소 (small -> footnotesize)
\footnotesize
% 2. 열(가로) 간격 및 행(세로) 간격 축소
\setlength{\tabcolsep}{3.5pt}
\renewcommand{\arraystretch}{0.95}
% 3. 표 전체 너비를 텍스트 너비의 100%에서 85% 수준으로 줄임
\resizebox{0.85\textwidth}{!}{%
\begin{tabular}{llrcccccc}
\toprule
\textbf{Group} & \textbf{Category} & \textbf{N} & \textbf{Greedy} & \textbf{Prompt-enh.} & \textbf{VCD} & \textbf{ICD} & \textbf{ALC} & \textbf{\Buzzy} \\
\midrule
\rowcolor[gray]{.95} \multicolumn{9}{l}{\textit{General Science}} \\
General & q-bio & 77 & 41.56 & \textbf{48.05} & 40.26 & 44.16 & 38.96 & 44.16 \\
General & econ & 79 & 62.03 & 50.63 & 54.43 & 55.70 & 51.90 & \textbf{63.29} \\
General & q-fin & 75 & 65.33 & 64.00 & 64.00 & 60.00 & 54.67 & \textbf{66.67} \\
General & stat & 61 & 72.13 & 68.85 & \textbf{73.77} & \textbf{73.77} & 70.49 & \textbf{73.77} \\
General & eess & 77 & 48.05 & \textbf{54.55} & 48.05 & 50.65 & 50.65 & 49.35 \\
General & math & 52 & 48.08 & 50.00 & 48.08 & 48.08 & \textbf{51.92} & \textbf{51.92} \\
General & physics & 79 & 75.95 & 68.35 & 75.95 & 77.22 & 67.09 & \textbf{78.48} \\
General & overall & 500 & 59.20 & 57.80 & 57.80 & 58.60 & 54.80 & \textbf{61.20} \\

\midrule
\rowcolor[gray]{.95} \multicolumn{9}{l}{\textit{Computer Science (CS)}} \\
CS & cross-list & 133 & 49.62 & 47.37 & 51.13 & 44.36 & 51.13 & \textbf{51.88} \\
CS & other cs & 132 & 46.97 & 48.48 & 52.27 & 49.24 & 48.48 & \textbf{54.55} \\
CS & cs.CR & 25 & 80.00 & 80.00 & \textbf{84.00} & 64.00 & 64.00 & 76.00 \\
CS & cs.DC & 25 & 44.00 & 44.00 & 48.00 & 44.00 & 48.00 & \textbf{56.00} \\
CS & cs.CV & 25 & 48.00 & 52.00 & \textbf{56.00} & 44.00 & 44.00 & 52.00 \\
CS & cs.RO & 25 & 52.00 & 40.00 & 48.00 & 44.00 & 52.00 & \textbf{56.00} \\
CS & cs.SE & 25 & 48.00 & 56.00 & 56.00 & \textbf{60.00} & 48.00 & \textbf{60.00} \\
CS & cs.NI & 25 & 64.00 & 60.00 & 60.00 & 44.00 & \textbf{68.00} & 60.00 \\
CS & cs.SY & 25 & \textbf{52.00} & 44.00 & 48.00 & 44.00 & 48.00 & 44.00 \\
CS & cs.LG & 25 & 56.00 & 56.00 & \textbf{64.00} & 52.00 & \textbf{64.00} & 60.00 \\
CS & cs.CL & 25 & \textbf{56.00} & \textbf{56.00} & \textbf{56.00} & 48.00 & \textbf{56.00} & \textbf{56.00} \\
CS & cs.AI & 10 & 20.00 & \textbf{30.00} & \textbf{30.00} & 20.00 & 20.00 & \textbf{30.00} \\
CS & overall & 500 & 51.00 & 50.40 & 54.00 & 47.40 & 51.40 & \textbf{54.80} \\

\midrule
\rowcolor[gray]{.95} Average & all & 1000 & 55.10 & 54.10 & 55.90 & 53.00 & 53.10 & \textbf{58.00} \\
\bottomrule
\end{tabular}%
}
\caption{Detailed category-wise accuracy (\%) on SciFIBench with the Qwen3.5-9B backbone. SciFIBench consists of two subsets: General Science and Computer Science (CS). The Average row reports the sample-weighted accuracy over all 1,000 examples.}
\label{tab:scifibench_category_appendix}
\end{table*}

\begin{table*}[t]
\centering
\small
\setlength{\tabcolsep}{4.5pt}
\renewcommand{\arraystretch}{1.05}
\resizebox{0.92\textwidth}{!}{%
\begin{tabular}{lrrrrrrr}
\toprule
\textbf{Category} & \textbf{N} & \textbf{Greedy} & \textbf{Prompt-enh.} & \textbf{VCD} & \textbf{ICD} & \textbf{ALC} & \textbf{\Buzzy} \\
\midrule
Biological sciences & 2062 & 44.23 & 42.48 & 48.93 & 45.25 & 46.02 & \textbf{51.16} \\
Physical sciences & 1039 & 49.95 & 49.18 & 50.05 & 48.22 & 47.93 & \textbf{53.61} \\
Health sciences & 330 & 49.09 & 46.97 & 50.30 & 47.27 & 48.48 & \textbf{53.03} \\
Earth and environmental sciences & 246 & 50.00 & 50.81 & 50.41 & 51.22 & 48.78 & \textbf{54.47} \\
Scientific community and society & 34 & 32.35 & 32.35 & 29.41 & \textbf{44.12} & 26.47 & 41.18 \\
\midrule
\rowcolor[gray]{.95} Average & 3711 & 46.54 & 45.22 & 49.29 & 46.65 & 46.78 & \textbf{52.14} \\
\bottomrule
\end{tabular}%
}
\caption{Category-wise accuracy (\%) on MMSci with the Qwen3.5-9B backbone. The Average row reports the sample-weighted accuracy over all 3,711 examples.}
\label{tab:mmsci_category_results}
\end{table*}

\begin{table*}[t]
\centering
\small
\setlength{\tabcolsep}{4.5pt}
\renewcommand{\arraystretch}{1.05}
\resizebox{\textwidth}{!}{%
\begin{tabular}{lrrrrrrr}
\toprule
\textbf{Category} & \textbf{N} & \textbf{Greedy} & \textbf{Prompt-enh.} & \textbf{VCD} & \textbf{ICD} & \textbf{ALC} & \textbf{\Buzzy} \\
\midrule
image\_scene & 407 & 91.9 & 94.3 & 93.6 & 88.9 & 92.1 & \textbf{95.8} \\
celebrity\_recognition & 396 & 91.4 & 91.7 & 91.2 & 86.9 & 91.7 & \textbf{92.9} \\
object\_localization & 315 & 50.2 & 50.5 & 53.0 & 48.6 & 49.2 & \textbf{54.3} \\
function\_reasoning & 304 & 80.3 & 81.2 & 79.6 & 76.6 & 81.2 & \textbf{83.2} \\
structuralized\_imagetext\_understanding & 282 & 86.9 & 86.5 & \textbf{87.2} & 85.1 & 85.8 & 86.5 \\
attribute\_recognition & 264 & 83.0 & \textbf{86.4} & 83.3 & 76.5 & 84.5 & 84.8 \\
physical\_property\_reasoning & 219 & 70.3 & 66.2 & 69.9 & 67.1 & \textbf{70.8} & 64.4 \\
action\_recognition & 215 & 90.2 & 92.1 & \textbf{94.4} & 88.4 & 88.4 & \textbf{94.4} \\
image\_style & 212 & 84.0 & 83.0 & 85.8 & 64.2 & 79.2 & \textbf{87.3} \\
image\_emotion & 200 & 76.0 & 78.5 & 80.0 & 71.5 & 74.0 & \textbf{81.5} \\
nature\_relation & 179 & 63.7 & 61.5 & \textbf{69.3} & 65.9 & 65.4 & 68.2 \\
spatial\_relationship & 177 & 52.5 & \textbf{55.9} & 50.8 & 53.1 & 50.8 & 50.3 \\
identity\_reasoning & 176 & 88.1 & 88.6 & 91.5 & 82.4 & 90.9 & \textbf{93.8} \\
social\_relation & 172 & 62.2 & 67.4 & 69.2 & 61.6 & 65.1 & \textbf{74.4} \\
ocr & 156 & 86.5 & \textbf{89.7} & \textbf{89.7} & 81.4 & 85.3 & \textbf{89.7} \\
image\_quality & 150 & 50.7 & 49.3 & 49.3 & 52.0 & \textbf{53.3} & 52.0 \\
attribute\_comparison & 141 & 66.7 & \textbf{70.2} & 67.4 & 64.5 & 66.7 & 68.8 \\
image\_topic & 140 & 89.3 & \textbf{93.6} & 91.4 & 90.0 & 89.3 & 92.1 \\
future\_prediction & 130 & 55.4 & 56.2 & 58.5 & 48.5 & 56.2 & \textbf{59.2} \\
physical\_relation & 94 & 63.8 & \textbf{66.0} & 62.8 & 60.6 & 64.9 & 62.8 \\
\midrule
\rowcolor[gray]{.95} Average & 4329 & 76.5 & 77.6 & 78.1 & 72.9 & 76.5 & \textbf{79.1} \\
\bottomrule
\end{tabular}%
}
\caption{Category-wise accuracy (\%) of Qwen3.5-9B on MMBench categories. The Average row reports sample-weighted accuracy over all examples.}
\label{tab:qwen35_9b_mmbench_category_breakdown}
\end{table*}

\begin{table*}[t]
\centering
\small
\setlength{\tabcolsep}{4.5pt}
\renewcommand{\arraystretch}{1.05}
\resizebox{\textwidth}{!}{%
\begin{tabular}{lrrrrrrr}
\toprule
\textbf{Category} & \textbf{N} & \textbf{Greedy} & \textbf{Prompt-enh.} & \textbf{VCD} & \textbf{ICD} & \textbf{ALC} & \textbf{\Buzzy} \\
\midrule
coarse perception & 250 & 57.2 & 55.6 & 54.0 & 49.2 & 55.6 & \textbf{60.4} \\
fine-grained perception & 250 & 31.6 & 31.2 & 30.4 & 30.4 & 28.8 & \textbf{35.2} \\
instance reasoning & 250 & 46.0 & 44.4 & 45.2 & 46.4 & 46.8 & \textbf{49.6} \\
logical reasoning & 250 & 34.4 & 34.4 & \textbf{38.0} & 33.6 & 36.0 & \textbf{38.0} \\
math & 250 & \textbf{45.6} & 44.4 & 44.4 & 44.8 & \textbf{45.6} & 43.6 \\
science \& technology & 250 & 31.2 & 32.8 & 33.2 & 34.8 & 30.0 & \textbf{35.6} \\
\midrule
\rowcolor[gray]{.95} Average & 1500 & 41.0 & 40.5 & 40.9 & 39.9 & 40.5 & \textbf{43.7} \\
\bottomrule
\end{tabular}%
}
\caption{Category-wise accuracy (\%) of Qwen3.5-9B on MMStar coarse categories. The Average row reports sample-weighted accuracy over all examples.}
\label{tab:qwen35_9b_mmstar_category_breakdown}
\end{table*}

\begin{table*}[t]
\centering
\small
\setlength{\tabcolsep}{4.5pt}
\renewcommand{\arraystretch}{1.05}
\resizebox{\textwidth}{!}{%
\begin{tabular}{lrrrrrrr}
\toprule
\textbf{Category} & \textbf{N} & \textbf{Greedy} & \textbf{Prompt-enh.} & \textbf{VCD} & \textbf{ICD} & \textbf{ALC} & \textbf{\Buzzy} \\
\midrule
image scene and topic & 141 & 44.0 & 39.7 & 37.6 & 37.6 & 41.8 & \textbf{44.7} \\
image emotion & 31 & 64.5 & 64.5 & 67.7 & 51.6 & 71.0 & \textbf{74.2} \\
image style \& quality & 78 & 78.2 & 80.8 & 78.2 & 69.2 & 74.4 & \textbf{83.3} \\
recognition & 118 & 35.6 & 35.6 & 37.3 & 34.7 & 32.2 & \textbf{41.5} \\
object counting & 92 & 23.9 & \textbf{25.0} & 19.6 & 20.7 & 22.8 & 23.9 \\
localization & 40 & 37.5 & 32.5 & 35.0 & 40.0 & 32.5 & \textbf{42.5} \\
cross-instance attribute reasoning & 89 & 47.2 & 40.4 & 40.4 & 44.9 & 46.1 & \textbf{50.6} \\
single-instance reasoning & 99 & 50.5 & 52.5 & 48.5 & 48.5 & 52.5 & \textbf{54.5} \\
cross-instance relation reasoning & 62 & 37.1 & 37.1 & \textbf{46.8} & 45.2 & 38.7 & 40.3 \\
common reasoning & 101 & 25.7 & 24.8 & \textbf{30.7} & 21.8 & \textbf{30.7} & 29.7 \\
diagram reasoning & 110 & 34.5 & 38.2 & 37.3 & 35.5 & 33.6 & \textbf{39.1} \\
code \& sequence reasoning & 39 & 56.4 & 48.7 & \textbf{59.0} & \textbf{59.0} & 56.4 & 56.4 \\
geometry & 116 & \textbf{51.7} & 48.3 & 48.3 & 50.0 & \textbf{51.7} & 47.4 \\
numeric commonsense and calculation & 48 & 39.6 & 37.5 & 43.8 & \textbf{45.8} & 41.7 & 39.6 \\
statistical reasoning & 86 & 40.7 & \textbf{43.0} & 39.5 & 37.2 & 39.5 & 40.7 \\
biology \& chemistry \& physics & 146 & 26.7 & 30.1 & 26.7 & \textbf{31.5} & 24.7 & \textbf{31.5} \\
electronics \& energy \& mechanical eng. & 46 & 56.5 & 52.2 & 58.7 & \textbf{60.9} & 56.5 & 56.5 \\
geography \& earth science \& agriculture & 58 & 22.4 & 24.1 & \textbf{29.3} & 22.4 & 22.4 & \textbf{29.3} \\
\midrule
\rowcolor[gray]{.95} Average & 1500 & 41.0 & 40.5 & 40.9 & 39.9 & 40.5 & \textbf{43.7} \\
\bottomrule
\end{tabular}%
}
\caption{Category-wise accuracy (\%) of Qwen3.5-9B on MMStar subcategories. The Average row reports sample-weighted accuracy over all examples.}
\label{tab:qwen35_9b_mmstar_l2_category_breakdown}
\end{table*}

\end{document}